\def\year{2022}\relax
\documentclass[letterpaper]{article} %
\usepackage{aaai22}  %
\usepackage{times}  %
\usepackage{helvet}  %
\usepackage{courier}  %
\usepackage[hyphens]{url}  %
\usepackage{graphicx} %
\usepackage{natbib}  %
\usepackage{caption} %
\DeclareCaptionStyle{ruled}{labelfont=normalfont,labelsep=colon,strut=off} %
\usepackage{algorithm}
\usepackage[utf8]{inputenc} %
\usepackage[T1]{fontenc}    %
\usepackage{hyperref}       %
\usepackage{url}            %
\usepackage{booktabs}       %
\usepackage{amsfonts}       %
\usepackage{nicefrac}       %
\usepackage{microtype}      %
\usepackage{amsfonts}       %
\usepackage{amssymb,amsmath,amsthm}
\usepackage{multirow}
\usepackage{algorithm}
\usepackage{algpseudocode}  %
\usepackage{graphicx}

\usepackage{lipsum}         %
\usepackage{xspace}         %
\usepackage{xargs}          %
\usepackage{mathtools}
\usepackage{bm}
\usepackage[dvipsnames]{xcolor} %
\usepackage{tabularx}       %
\usepackage{wrapfig}        %
\usepackage{tikz}           %
\usepackage{ifthen}
\usepackage{enumitem}%
\usepackage[utf8]{inputenc} %
\usepackage[T1]{fontenc}    %
\usepackage[]{hyperref}       %
\usepackage{url}            %
\usepackage{booktabs}       %
\usepackage{amsfonts}       %
\usepackage{amssymb,amsmath,amsthm}
\usepackage{nicefrac}       %
\usepackage{microtype}      %
\usepackage{algorithm}
\usepackage{algpseudocode}  %
\usepackage{graphicx}
\usepackage{tikzit}
\usepackage{subcaption}
\usepackage{lipsum}         %
\usepackage{xspace}         %
\usepackage{xargs}          %
\usepackage{mathtools}
\usepackage{bm}
\usepackage[dvipsnames]{xcolor} %
\usepackage{tabularx}       %
\usepackage{wrapfig}        %

\usepackage{tikz}           %
\usepackage{ifthen}
\usepackage{enumitem}

\usetikzlibrary{patterns}
\usetikzlibrary{positioning}

\expandafter\def\expandafter\normalsize\expandafter{%
    \normalsize
    \setlength\belowdisplayskip{6pt}
    \setlength\belowdisplayshortskip{4pt}
}

\makeatletter
\newcommand{\currentfontsize}{\f@size pt}
\makeatother

\usepackage[compact]{titlesec} 
\titlespacing*{\subsection}{6pt}{3pt}{1pt}

\newcommandx{\customComment}[3]{\textcolor{#2}{\textsl{#1: #3}}}
\newcommandx{\customTodo}[3]{\textcolor{#2}{\textsl{#1: #3}}}

\newcommandx{\Martin}[1]{\customComment{Martin}{brown}{#1}}
\newcommandx{\NeedsRef}[1]{\textcolor{red}{~(ref)}}

\DeclarePairedDelimiterX{\lin}[2]{\langle}{\rangle}{#1, #2}

\DeclarePairedDelimiterX{\abs}[1]{\lvert}{\rvert}{#1}
\DeclarePairedDelimiterX{\norm}[1]{\lVert}{\rVert}{#1}
\DeclarePairedDelimiterX{\cbr}[1]{\{}{\}}{#1} %
\DeclarePairedDelimiterX{\rbr}[1]{(}{)}{#1} %
\DeclarePairedDelimiterX{\sbr}[1]{[}{]}{#1} %

\providecommand{\real}{\mathbb{R}} %

\DeclareMathOperator{\expect}{\mathbb{E}}

\DeclareMathOperator{\sgn}{sign}
\makeatletter
\def\sign{\@ifnextchar*{\@sgnargscaled}{\@ifnextchar[{\sgnargscaleas}{\@ifnextchar{\bgroup}{\@sgnarg}{\sgn} }}}
\def\@sgnarg#1{\sgn\rbr{#1}}
\def\@sgnargscaled#1{\sgn\rbr*{#1}}
\def\@sgnargscaleas[#1]#2{\sgn\rbr[#1]{#2}}
\makeatother

\providecommand{\0}{\mathbf{0}}

\providecommand{\uu}{\mathbf{u}}

\providecommand{\vv}{\mathbf{v}}

\providecommand{\xx}{\mathbf{x}}
\providecommand{\yy}{\mathbf{y}}

\providecommand{\mX}{\mathbf{X}}

\providecommand{\cO}{\mathcal{O}}

\newtheorem{theorem}{Theorem}

\newtheorem{lemma}{Lemma}

\providecommand{\rb[1]}{\left(#1\right)}

\newcommand{\E}{\mathbb{E}}

\newcommand{\xt}[1]{\xx^{(#1)}}
\newcommand{\Ea}[1]{\E\left[#1\right]}
\newcommand{\Eb}[2]{\E_{#1}\left[#2\right]}

\newcommand{\og}{\overline{g}}

\tikzstyle{server}=[fill={rgb,255: red,64; green,64; blue,64}, draw=black, shape=circle]
\tikzstyle{client}=[fill={rgb,255: red,55; green,126; blue,184}, draw=black, shape=circle]
\tikzstyle{optima}=[fill={rgb,255: red,55; green,126; blue,184}, draw=black, shape=rectangle]
\tikzstyle{globalopt}=[fill={rgb,255: red,255; green,128; blue,0}, draw=black, shape=rectangle]
\tikzstyle{pseudoserver}=[fill={rgb,255: red,191; green,191; blue,191}, draw={rgb,255: red,128; green,128; blue,128}, shape=circle]
\tikzstyle{sgdnode}=[fill=white, draw=black, shape=circle]
\tikzstyle{client2}=[fill={rgb,255: red,77; green,175; blue,74}, draw=black, shape=circle]
\tikzstyle{clientopt}=[fill={rgb,255: red,77; green,175; blue,74}, draw=black, shape=rectangle]
\tikzstyle{averageopt}=[fill={rgb,255: red,191; green,191; blue,191}, draw={rgb,255: red,128; green,128; blue,128}, shape=rectangle]
\tikzstyle{gradient}=[dashed]

\tikzstyle{server update}=[->, draw={rgb,255: red,77; green,175; blue,74}]
\tikzstyle{client update}=[draw={rgb,255: red,55; green,126; blue,184}, ->]
\tikzstyle{update}=[draw=black, dashed, -]
\tikzstyle{sgd}=[draw={rgb,255: red,65; green,61; blue,58}, ->]
\tikzstyle{correction}=[draw={rgb,255: red,228; green,26; blue,28}, ->]
\tikzstyle{midpoint}=[-, draw={rgb,255: red,191; green,191; blue,191}]

\makeatletter
\newcommand{\myitem}[1]{%
\item[\textbf{(#1)}]\protected@edef\@currentlabel{#1}%
}
\makeatother

  \providecommand{\real}{\mathbb{R}} %

  \makeatletter
  \def\sign{\@ifnextchar*{\@sgnargscaled}{\@ifnextchar[{\sgnargscaleas}{\@ifnextchar{\bgroup}{\@sgnarg}{\sgn} }}}
  \def\@sgnarg#1{\sgn\rbr{#1}}
  \def\@sgnargscaled#1{\sgn\rbr*{#1}}
  \def\@sgnargscaleas[#1]#2{\sgn\rbr[#1]{#2}}
  \makeatother

  \providecommand{\0}{\bm{0}}

  \providecommand{\uu}{\bm{u}}
  \renewcommand{\vv}{\bm{v}}
  
  \providecommand{\xx}{\bm{x}}
  \providecommand{\yy}{\bm{y}}

  \providecommand{\mX}{\bm{X}}

  \providecommand{\cO}{\mathcal{O}}

\newcommand{\ignore}[1]{}

\definecolor{color1}{RGB}{228,26,28}
\definecolor{color2}{RGB}{55,126,184}
\definecolor{color3}{RGB}{77,175,74}
\definecolor{color4}{RGB}{152,78,163}
\definecolor{color5}{RGB}{255,127,0}


\usetikzlibrary{fit,calc}

\colorlet{client}{red!40}
\newcommand{\speedup}[1]{{\color{gray}(\ifdim #1 pt > 0.3pt #1\else $< #1$\fi{}$\times$)}}

\newcommandx*\prm[2][1=x]{\mX_{#2}^{\ifthenelse{ \equal{#1}{x} }{}{(#1)}}}

\usepackage{newfloat}
\usepackage{listings}

\floatstyle{ruled}
\newfloat{listing}{tb}{lst}{}
\floatname{listing}{Listing}
\title{Implicit Gradient Alignment in Distributed and Federated Learning}
\author {
    Yatin Dandi\thanks{Denotes equal contribution.},\textsuperscript{\rm 1} \textsuperscript{\rm 2}
    Luis Barba\footnotemark[1], \textsuperscript{\rm 2}
    Martin Jaggi \textsuperscript{\rm 2}
}
\affiliations {
    \textsuperscript{\rm 1} IIT Kanpur, India\\
    \textsuperscript{\rm 2} EPFL, Switzerland\\
    yatind@iitk.ac.in,
    luis.barbaflores@epfl.ch,
    martin.jaggi@epfl.ch
}

\usepackage{bibentry}

\begin{document}
\maketitle
\begin{abstract}
A major obstacle to achieving global convergence in distributed and federated learning is the misalignment of gradients across clients, or mini-batches due to heterogeneity and stochasticity of the distributed data. In this work, we show that data heterogeneity can in fact be exploited to improve generalization performance through implicit regularization.
One way to alleviate the effects of heterogeneity is to encourage the alignment of gradients across different clients throughout training. Our analysis reveals that this goal can be accomplished by utilizing the right optimization method that replicates the implicit regularization effect of SGD, leading to gradient alignment as well as improvements in test accuracies.
Since the existence of this regularization in SGD completely relies on the sequential use of different mini-batches during training, it is inherently absent when training with large mini-batches.
To obtain the generalization benefits of this regularization while increasing parallelism, we propose a novel GradAlign algorithm that induces the same implicit regularization while allowing the use of arbitrarily large batches in each update. We experimentally validate the benefits of our algorithm in different distributed and federated learning settings.
\end{abstract}

\section{Introduction}\label{sec:intro}
In this paper we focus on sum structured optimization of the form $f(\xx):= \frac{1}{n}\sum_{i=1}^n f_i(\xx)$, where each $f_i$ is a different function representing the loss function of either distinct data points, mini-batches or clients.
To prove convergence, many assumptions over the $f_i$'s have been studied. For example, one may assume fixed bounds on the variance or dissimilarity of gradients across different~$f_i$.  However, in practice, for non-convex optimization problems such as deep neural networks, the dissimilarity across gradients is likely to vary across different values of $\xx$.
 We instead argue that to obtain optimal generalization performance, it is desirable to not only converge to a solution that minimizes the mean loss $f(\xx)$, but also encourage convergence to regions with reduced gradient dissimilarity.

We propose to achieve convergence to such solutions by aligning the gradients across different~$f_i$. 
To this end, we introduce a regularizer $r(\xx) = \frac{1}{2n}\sum_{i=1}^n\norm{\nabla f_i(\xx) - \nabla f(\xx)}^2$ measuring the variance of gradients across the mini-batches, whose minimization leads to the alignment of different gradients.
As demonstrated recently by \citet{smith2021on}, stochastic gradient descent (SGD) \citep{robbins1951sgd} already contains an \emph{implicit} regularization effect over gradient descent (GD) corresponding to the minimization of $r(\xx)$, when comparing updates over an entire epoch.
Our analysis applicable to arbitrary sequences of SGD steps further reveals that the optimization trajectory followed by SGD can be approximated through gradient descent on the \emph{surrogate function} $\hat{f}(\xx) := f(\xx) + \lambda r(\xx)$ with the strength of the regularization being controlled by the step size.
This motivates us to devise new algorithms tailored to implicitly minimize this surrogate function~$\hat{f}(\xx)$.

While control variates-based variance reduction techniques can effectively reduce the variance across different updates \citep{DBLP:conf/nips/Johnson013}, they do not directly promote variance reduction through the alignment of different $f_i$'s gradients for the current iterate, i.e., such methods do not encourage the decrease of $r(\xx)$ throughout training.
A small variance of gradients across mini-batches, i.e., small~$r(\xx)$, corresponds to the alignment of gradients for different data points. Such alignment can benefit generalization throughout training, since large gradient alignment across datapoints implies that gradient updates on $f_i$ corresponding to empirical risk on a subset of the data may reduce the loss for a much larger number of data points, even outside the training set. A similar observation was recently  utilized to improve transfer in error reduction across datapoints in meta-learning \citep{nichol2018firstorder}. 
The gradient alignment in SGD arises due to its sequential nature and the use of small mini-batches, which together induce dependencies between successive updates contributing to the implicit minimization of $r(\xx)$.
These effects, however, decrease as the mini-batch size is increased, since the variance across mini-batches diminishes.
This imposes a trade-off between using large mini-batches per update and obtaining gradient alignment and hence better generalization.
A similar trade-off has been observed empirically~\citep{DBLP:conf/iclr/KeskarMNST17,pmlr-v80-ma18a,pmlr-v84-yin18a}, where using larger mini-batches has been shown to worsen the generalization performance. 

We argue that the utilization of gradient alignment to improve generalization can be especially beneficial in distributed and federated learning. In datacenter distributed learning \citep{goyal2018accurate,NIPS2012_6aca9700}, where the primary bottleneck is the computation of gradients instead of communication, \citep{MAL-083}, it is desirable to exploit the available parallelism to the maximum extent, without losing the benefits of sequential updates on small mini-batches provided by SGD. Our proposed algorithm, GradAlign, achieves this by aligning the gradients across clients through implicit regularization. 

In a federated setting \citep{2016federated}, where multiple updates for each client are required to reduce the communication cost, data dissimilarity among clients plays an especially important role. One common approach to obtain the regularization benefits of SGD in federated learning is to run SGD on small mini-batches in parallel on separate clients, each with a different subset of the data, while periodically averaging the iterates to obtain global updates (FedAvg~\citep{pmlr-v54-mcmahan17a}). 
However, the local nature of optimization in each client, prevents gradient alignment across mini-batches corresponding to different clients.
Such gradient alignment across clients is particularly desirable in the presence of data heterogeneity across clients
where the convergence of Federated Averaging is hindered due to the phenomenon of ``client drift.''
\citep{pmlr-v119-karimireddy20a}, corresponding to the deviation of local updates for each client from the gradient of the global objective. Thus gradient alignment across clients in federated learning, analogous to the gradient alignment across mini-batches in SGD, would not only improve the test accuracy upon convergence, but also minimize the client drift in the presence of heterogeneity.
To achieve this, we design a novel algorithm Federated Gradient Alignment (\emph{FedGA}), that replicates the implicit regularization effect of SGD by promoting inter-client gradient alignment. We further derive the existence of a similar regularization effect in a recently proposed algorithm, SCAFFOLD \citep{pmlr-v119-karimireddy20a}, albeit without the ability to fine-tune the regularization coefficient. Our main contributions are thus as follows:
\begin{enumerate}
\item We design a novel algorithm GradAlign that replicates the regularization effect of a sequence of SGD steps while allowing the use of the entire set of mini-batches for each update.
\item We extend GradAlign to the federated learning setting as FedGA, and derive the existence of the implicit inter-client gradient alignment regularizer $r(\xx)$ for FedGA as well as for SCAFFOLD.
\item We derive sufficient conditions under which GradAlign causes a decrease in the explicitly regularized objective $\hat{f}(\xx)$. 
\item We empirically demonstrate that FedGA achieves better generalization than both FedAvg~\citep{pmlr-v54-mcmahan17a} and SCAFFOLD~\citep{pmlr-v119-karimireddy20a}.
\end{enumerate}

\section{Related Work}\label{sec:related work}

The relationship between the similarity of gradients and generalization has been explored in several recent works \citep{Chatterjee2020Coherent,chatterjee2020making,fort2020stiffness}. Our work strengthens the empirical findings in these papers and provides a mechanism to extend the benefits of gradient alignment to distributed and federated learning settings.

The generalization benefits of gradient alignment can also be interpreted through the lens of Neural Tangent Kernel \citep{NEURIPS2018_5a4be1fa}: the loss $l(\xx)$ at a test point $\xx$ and the prediction $y$ decreases as $\nabla l(y)^\top \frac{1}{n}\sum_{i=1}^n K(x, x_i)\nabla l_i(y_i)$, where $x_i, y_i$ correspond to training points, $K(x, x_i)$ represents the inner product between the output's gradient at test point $x$ and training point $x_i$ and $\nabla l(y), \nabla l_i(y_i)$ denote the gradient of the loss w.r.t the outputs at the corresponding points.
Thus, test points with high gradient similarity lead to a larger decrease in their loss.
Our work corroborates the recent empirical findings in \citep{lin2020extrapolation}, where the use of extrapolation for large batch SGD leads to significant gains in generalization performance. While \citet{lin2020extrapolation} attributed the improved generalization to smoothening of the landscape due to extrapolation, our analysis and results provide a novel perspective to the benefits of displacement through implicit regularization.

The generalization benefits of SGD have been analyzed through several related perspectives such as Stochastic Differential Equations (SDEs) \citep{chaudhari2018stochastic,jastrzebski2018factors}, Bayesian analysis \citep{l.2018a,mandtsgd} and flatness of minima \citep{YaoHessian,DBLP:conf/iclr/KeskarMNST17}, which has been challenged by \citet{pmlr-v70-dinh17b}. Unlike these works, the implicit regularization perspectives in \citet{barrett2021implicit} and our work directly describe a modified objective upon which gradient flow and gradient descent respectively approximate the updates of SGD. Moreover, our analysis incorporates the effects of finite step sizes, whereas the SDE-based analysis relies on infinitesimal learning rates.

The existence of shared optima in sum structured optimization has previously been analyzed in the context of a strongly convex objective, where the strong growth condition \citep{schmidt2013fast} implies the existence of a shared optimum and linear convergence for both deterministic and stochastic gradient descent. However, for general non-convex objectives having multiple local minima, it is desirable to encourage convergence to the set of minima to the ones being nearly optimal for all the components $f_i$ without sacrificing the ability to use large amounts of data for each update.

A large number of works have attempted to modify the FedAvg algorithm to improve convergence rates and minimize client drift.
Our implicit regularization can easily be incorporated into the various modifications of FedAvg such as FedProx \citep{fedprox}, FedDyn \citep{feddyn}, FedAvgM \citep{fedavgm}, FedAdam \citep{reddi2021adaptive}, etc. by introducing the displacements used in our algorithms into the local gradient updates used in these algorithms. Comparisons against FedProx in the Experiments section further verify the utility of our approach as a standalone modification in heterogeneous as well as i.i.d federated learning settings.

\section{Setup}\label{sec:setup}
We consider the standard setting of empirical risk minimization with parameters $\xx$, represented as a sum\vspace{-3mm}
\[
    \min_{\xx \in \real^d} \cbr[\Big]{ f(\xx)
:= \frac{1}{n}\sum_{i=1}^n f_i(\xx)}\,,
\]
where the function $f_i$ denotes the empirical risks on the $ i_{th}$ subset of the training data. Here the subsets correspond to different mini-batches, clients, or clients depending on the application. We further define the regularizer \vspace{-2mm}
\[
r(\xx) = \frac{1}{2n}\sum_{i=1}^n\norm{\nabla f_i(\xx) - \nabla f(\xx)}^2.
\]
Here $r(\xx)$ represents $\frac{1}{2}$ times the trace of the covariance matrix for the mini-batch gradients. The gradient of $r(\xx)$ is then given by:\vspace{-2mm}
      \begin{align*}
          \nabla r(\xx) = \frac{1}{n}\sum_{i=1}^{n}\left(\nabla^2 f_i(\xx)-\nabla^2 f(\xx)\right) (\nabla f_i(\xx)-\nabla f(\xx) ).
      \end{align*}

\section{Analysis and Proposed Algorithms}\label{sec:main}

A key component in all our subsequent analysis is the expression for the gradient of $f_i$ at a point obtained after applying a displacement $\vv_{\xx}$ to a given point $\xx$, i.e., $\nabla f_i(\xx + \vv_{\xx})$. By applying Taylor's theorem to each component of $\nabla f_i$, we obtain the following expression (see Appendix \ref{app:proofs}):
\begin{lemma}\label{lemma:displacement}
If $f_i$ has Lipschitz  Hessian, i.e., $\norm{\nabla^2 f_i(\xx)-\nabla^2 f_i(\yy)}_2 \leq \rho\norm{\xx-\yy}$ for some $\rho > 0$, then
\begin{equation}
\label{eq:Taylor}
    \nabla f_i (\xx+\vv_{\xx}) =  \nabla f_i (\xx) + \nabla^2 f_i (\xx)\vv_{\xx} + \cO(\norm{\vv_{\xx}}^2).
\end{equation}
\end{lemma}

For instance, when $\vv_{\xx} = -\alpha\nabla f_i(\xx)$, we have:
\begin{equation}\label{eq:example}
    \nabla f_i (\xx-\alpha\nabla f_j(\xx)) = \nabla f_i (\xx) - \alpha\nabla^2 f_i (\xx)\nabla f_j(\xx)+ \cO(\alpha^2)
\end{equation}

\subsection{Implicit Gradient Alignment}
Suppose that, given two minibatches corresponding to objectives $f_i, f_j$, we optimize the parameters $\xx$ by first updating in the direction of the negative gradient of say $f_i$ and then compute the gradient with respect to the new mini-batch, say $f_j$, i.e., we utilize $\nabla f_j(\xx -\alpha\nabla f_i(\xx))$ for the second update. 
From Equation \ref{eq:example}, we observe that, when the order of gradient steps on $f_i$ and $f_j$, is random, second-order term due to displacement (Lemma \ref{lemma:displacement}) in expectation equals $-\frac{\alpha}{2}\rb{\nabla^2 f_i (\xx)\nabla f_j(\xx)+\nabla^2 f_j (\xx)\nabla f_i(\xx)} = - \frac{\alpha}{2}\nabla\rb{{\nabla f_i (\xx)}^\top{\nabla f_j(\xx)}}$. \textbf{Thus for two given minibatches, i and j, sequential SGD steps in random order lead to implicit maximization of the inner product of the corresponding gradients. We refer to this phenomenon of alignment of gradients across mini-batches as ``Implicit Gradient Alignment''.}
In the next section, we generalize this argument to arbitrary sequences of minibatches.
\subsection{SGD over K Sequential Steps}\label{sec:k sequential steps}

Recall that SGD computes gradients with respect to randomly sampled mini-batches in each round. As explained above, such sequential updates on different mini-batches implicitly align the gradients corresponding to different minibatches. 
We make this precise by deriving the implicit regularization in SGD for a sequence of $K$ steps under SGD. 
A similar regularization term was derived by \citet{smith2021on} in the context of backward error analysis for the case of a sequence corresponding to non-overlapping batches covering the entire dataset.  
They derived a surrogate loss function upon which gradient flow approximates the path followed by SGD when optimizing the original loss function $f$.
Since continuous-time gradient flow is unusable in practice, we instead aim to derive a surrogate loss function $\hat{f}$ where a large batch gradient descent algorithm on this surrogate loss would approximate the path followed by SGD when optimizing $f$.

Moreover, our analysis applies to arbitrary $K$ and any sampling procedure symmetric w.r.t time, i.e, we only assume that for any sequence of $K$ mini-batches $A = \{a_i\}^{K}_{i=1}$, the corresponding reverse sequence $A_{-1} = \{a_{K+1-i}\}^{K}_{i=1}$ has the same probability. This allows us to conveniently evaluate the average effect of SGD for a particular sequence over all possible re-orderings of the sequence. Note that this assumption is valid both when sampling with and without replacement from any arbitrary distribution over mini-batches.

While each gradient update in SGD is an unbiased estimate of the full gradient, the cumulative effect of multiple updates on randomly sampled mini-batches can differ from the minimization of the original objective, as illustrated through Equation~(\ref{eq:example}).
To isolate the effect of sequential updates on particular sequences of sampled mini-batches, we compare the steps taken by SGD against the same number of steps using GD on the sample mean of the sequence's objective.
We denote the gradient and Hessian for mini-batch $a_i$ by $\nabla f_{a_i}(\xx)$ and $\nabla^2 f_{a_i}(\xx)$ respectively while $\nabla f_{A}(\xx), \nabla^2 f_A(\xx)$ denote the mean gradient and Hessian for the entire sequence $A$.
By applying Lemma \ref{eq:Taylor} to each gradient step, we obtain the following result (proof in the Appendix \ref{app:proofs}):
\begin{theorem}\label{thm:SGD}
Conditioned on the (multi)set of mini-batches in a randomly sampled sequence $A$ of length $K$, the expected difference between the parameters reached after $K$ steps of SGD using the corresponding mini-batches in $A$ and $K$  steps of GD with step size $\alpha$ on the mean objective $f_A(\xx) = \frac{1}{K}\sum_{i=1}^K f_{a_i}(\xx)$, both starting from the same initial parameters $\xx$ is  given by:\vspace{-1mm}
\begin{align}
    &\Ea{\xx_{SGD,A}-\xx_{GD,A}}\nonumber\\=&
    -\frac{\alpha^2}{2} \Big( \sum_{i=1}^{K} (\nabla^2 f_{a_i}(\xx) \rb{\nabla f_{a_i}(\xx) - \nabla f_{A}(\xx)} \Big) + \cO(\alpha^3)\label{eq:displacement}\\
    & -\frac{K\alpha^2}{2}{\nabla r_A(\xx)} + \cO(\alpha^3) \label{eq:regularization}
\end{align}
\end{theorem}\vspace{-2mm}
where, analogous to Section~\ref{sec:setup}, we define $r_A(\xx) = \frac{1}{2K}\big(\sum_{i=1}^{K} \|\nabla f_{a_i}(\xx)-\nabla f_{A}(\xx)\|^2\big)$. 
For the particular case of a sequence covering an entire epoch, i.e. $K=n$ and sampling without replacement, we recover the implicit regularization over gradient descent derived by \citet{smith2021on}.
The above results imply that $K$ steps of SGD not only optimize the original objective function analogous to GD, but additionally move the parameters opposite to the gradient of 
$r_A(\xx)$
Thus, SGD implicitly minimizes $r_A(\xx)$ along with the original objective, which leads us to call the latter term an \emph{implicit regularizer}. 
As we show in the Appendix \ref{app:proofs}, the net displacement of SGD in Equation \eqref{eq:regularization} can be approximated by $K$ gradient descent steps on the mean objective regularized by $\frac{\alpha}{2}r_A(\xx)$. Thus optimizing the regularized objective can allow us to utilize $K$ times more data for each update, while still approximating the trajectory followed by SGD.
This is in contrast to the linear scaling rule discussed in \citet{goyal2018accurate}, which aims to approximate the sequence of $K$ SGD steps with a single GD step with a step size scaled by $K$. However, such linear scaling only approximates the first-order gradient terms in the sequence, ignoring the implicit gradient alignment. We discuss this further in Appendix \ref{app:linear}, and analyze a linearly scaled approximation of SGD that incorporates implicit gradient alignment. A crucial advantage of approximating SGD using the same number of gradient steps and step size is that it allows the use of larger total batch sizes, whereas linear scaling is only effective for batch sizes much smaller than the total training set size \citep{shallue2019batch}.

We observe that the term corresponding to the Hessian for the mini-batch~$a_i$ in Equation~\eqref{eq:displacement} can be obtained using as the product of the Hessian and the vector $\vv_\xx = -\frac{\alpha}{2}\rb{\nabla f_{A}(\xx)-\nabla f_{a_i}(\xx)}$.
Thus utilizing the right vector for each mini-batch allows us to approximate the regularization effect of SGD.
We further observe that Lemma~\ref{lemma:displacement} provides an efficient method for obtaining the Hessian-vector product by computing the gradient of $f_{a_i}$ on the point $\xx$ displayed by $\vv_\xx$, eliminating the time and memory overhead of explicit Hessian-gradient vector computation.
Moreover, as we illustrate in section \ref{subsec:federated}, the displacement-based formulation allows the utilization of gradient alignment in federated settings with multiple (K>1) local steps, without additional communication, computation time, or memory overhead for each of the local steps.
In the subsequent sections, we utilize these observations to design algorithms for distributed and federated learning that replicate the regularization effect of SGD while allowing parallelism for the use of arbitrarily large batches, overcoming the generalization failure of traditional large-batch training \citep{shallue2019batch}.

\begin{figure*}[t]
\centering\vspace{-1em}
\includegraphics[width=0.9\textwidth]{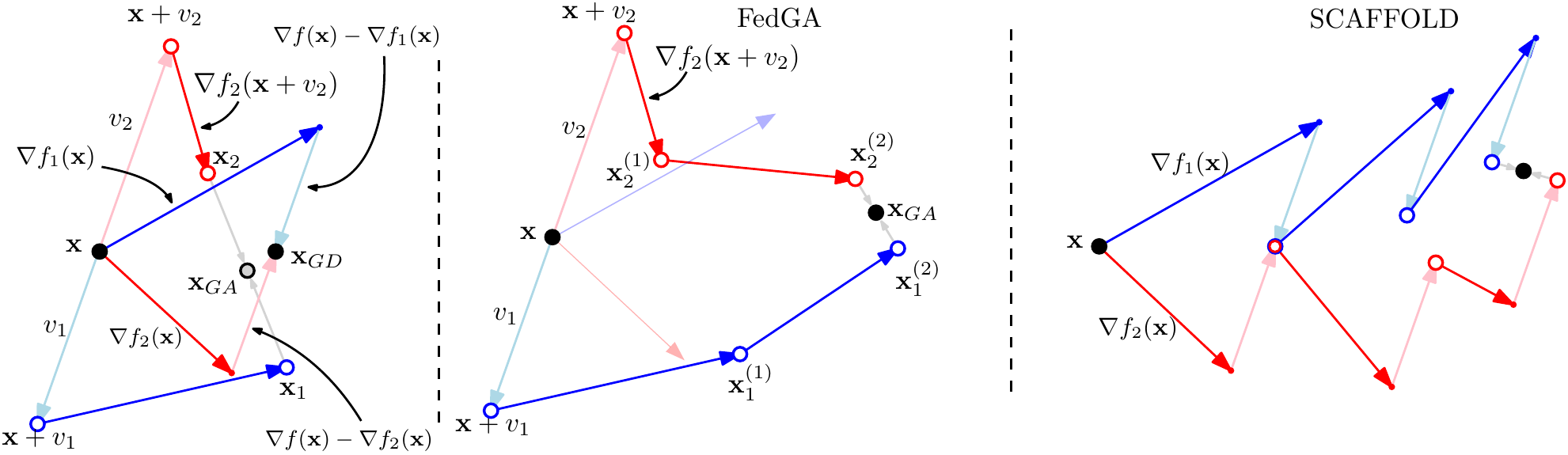}\vspace{-2mm}
\caption{\small Left: Depiction of one round of GD against one round of GradAlign (equivalent to one round of FedGA with $K=1$, see Appendix~\ref{app:FedGA}) along with the computation of the displacements $\vv_i = -\beta(\nabla{f}(\xx) - \nabla{f_i}(\xx))$. Middle: Schematic depiction of one round of FedGA consisting of $K=2$ steps. After the initial displacement of $\xx$, the algorithm follows $K$ local updates.  Right: Schematic depiction of one round of SCAFFOLD where the displacement is applied after each local update.}\vspace{-3mm}
\label{fig:FedGA_vs_SCAFFOLD}
\end{figure*}

\subsection{Gradient Alignment under Parallel Computations}\label{section:GradAlign}
The analysis in the previous section revealed that sequential updates on a randomly sampled set of mini-batches not only minimize the mean sampled objective but also the variance of gradients across the sampled mini-batches. We aim to replicate this effect while allowing the use of parallelism across mini-batches. Through Equation \eqref{eq:displacement} and Lemma \ref{lemma:displacement}, we observed that the source of gradient alignment in the sequential updates for SGD is the evaluation of the gradient of a mini-batch $i$ after an additional displacement in the direction of $-\rb{\nabla f(\xx)-\nabla f_i(\xx)}$. Thus we can replicate the gradient alignment of SGD by utilizing gradients for each mini-batch $i$ computed after an initial displacement $\vv_i(x) = -\beta\rb{\nabla f(\xx)- \nabla f_i(\xx)}$. This ensures that the vector multiplying $\nabla^2 f_i(\xx)$ due to displacement (Lemma~\ref{lemma:displacement}) matches the corresponding vector in the negative gradient of $\beta r(\xx) = \beta \frac{1}{2n}\sum_{i=1}^n\norm{\nabla f_i(\xx) - \nabla f(\xx)}^2$. Moreover, unlike SGD, the step size for the displacement $\beta$ can differ from $\frac{\alpha}{2}$, enabling the fine-tuning of the regularization coefficient. 
We refer to the resulting Algorithm~\ref{alg:GA} as GradAlign (GA).
\begin{algorithm}
  \caption{GradAlign (GA)}%
  \label{alg:GA}
  \begin{algorithmic}[1]{
      \State Learning rate $\alpha$, initial model parameters :$\xx$
      \While{not done}
        \State{$\nabla f(\xx) \gets \frac{1}{n}\sum_{i=1}^{n}\nabla f_i(\xx)$} 
        \Comment{Obtain the full gradient by computing the mini-batch gradients in parallel}
        \For{mini-batches $i$ in $[1,\cdots,n]$ in parallel}
          \State{Obtain the displacement for the $i_{th}$ minibatch as $\vv_i \gets \rb{\nabla f(\xx)- \nabla f_i(\xx)}$}
          \State{$\xx_i \gets \xx -  \alpha\nabla{f_i}(\xx-\beta \vv_i)$}
          \Comment{Obtain gradient after displacement}
          \EndFor
      \State{$\xx \gets \frac{1}{n}\sum_{i=1}^{n}\xx_i$}
      \EndWhile
      }
    \end{algorithmic}
\end{algorithm}
\begin{theorem}\label{thm:GA}
The difference between the parameters reached by one step of GradAlign with step size $\alpha$ and displacement~$\beta$ and gradient descent objective starting from the initial parameters $\xx$ is given by\vspace{-1mm}
\begin{align*}
    \xx_{GA} - \xx_{GD} &= -\frac{\alpha\beta}{2n}\nabla_{\xx}\Big(\sum_{i=1}^{n} \|\nabla f_i(\xx)-\nabla f(\xx)\|^2\Big)\\ &+\cO(\alpha\beta^2).
\end{align*}
\vspace{-3mm}
\end{theorem}

\textbf{Descent Condition.}
Since the displacement step size $\beta$ controls the strength of regularization as well as the error in approximating the gradient of the regularized objective, it is imperative to know if there exists a suitable range of $\beta$ under which GradAlign causes a decrease in the surrogate objective $\hat{f}(\xx) = f(\xx) + \beta r(\xx)$.
We prove that unless the algorithm is at a point that is simultaneously critical for $f(\xx)$ as well as $r(\xx)$, for sufficiently small step and displacement sizes, each step of FedGA causes a decrease in $\hat{f}(\xx)$. This lends credence to the use of GradAlign to ensure convergence to shared optima in distributed settings for general smooth non-convex objectives. The proof of the theorem, the justifications for the assumptions, and the consequences for convergence, are provided in the Appendix~\ref{app:descent condition}.

\begin{theorem}\label{thm:descent}
Assuming $L_1$-smoothness of $f(\xx)$, $L_2$-smoothness of $r(\xx)$, and Lipschitzness of Hessians, for $\xt{t}$ satisfying at least one of $\nabla f(\xt{t})\neq \0$ or $\nabla r(\xt{t}) \neq \0$, $\exists \beta > 0$ such that updating $\xt{t}$ using GradAlign with step size $\alpha < \frac{1}{2L_1}$ and displacement~$\beta$ results in updated parameters $\xt{t+1}$ satisfying $\hat{f}(\xt{t+1}) - \hat{f}(\xt{t}) < 0$.
\end{theorem}
While the above theorem suggests the possibility of requiring adaptation of the displacement step size with time, in practice, we found that a constant step size is sufficient to achieve significant gains in test accuracy. We hypothesize that this is due to the decrease in variance across mini-batch gradients over time, which balances the effect of the decrease in the gradient norm.

\subsection{Federated Learning}\label{subsec:federated}

In the presence of large communication costs across clients, it is desirable to allow multiple local updates for each client before each round of communication.
Such an approach is known in the literature as Federated Averaging (FedAvg) \citep{mcmahan2017communication} or local SGD, where each round involves $K >1$ updates on local objectives corresponding to the loss of randomly sampled clients. In the case of identical data distributions across clients, parts of the generalization benefits of SGD readily appear in FedAvg due to the sequential local update steps within each client \citep{parallelSGD}, leading to significant gains in test accuracies over gradient descent on large batches \citep{Lin2020Don't,pmlr-v119-woodworth20a}. However, as we prove in the appendix \ref{app:proofs}, local SGD steps lead to gradient alignment only across mini-batches within the same client. 
We argue that extending FedAvg to allow implicit gradient alignment \emph{across} clients is desirable for two major reasons.
First, similar to SGD and GradAlign, implicit regularization through the minimization of inter-client variance of the gradients is expected to improve generalization performance by encouraging convergence to shared optima across the different clients' objectives. Moreover, gradient alignment across clients crucially minimizes the effects of ``client drift'', where the presence of the heterogeneity in the data distributions across clients can cause each client's iterates to deviate from the optimization trajectory of the global objective significantly \citep{pmlr-v119-karimireddy20a}.

We consider a federated learning setup corresponding to the minimization of the average loss over~$n$ clients w.r.t. parameters $\xx$. For simplicity, we assume that all the $n$ clients are sampled in each round. 
We extend the GradAlign algorithm to the federated setting by computing the local updates for each client $i$ using the gradients obtained after an initial additive displacement $\vv_i(x) = -\beta\rb{\nabla f(\xx)- \nabla f_i(\xx)}$ obtained at the beginning of each round. Since the displacement for each client remains constant throughout a round, the displacement step $\vv_i$ needs to be applied only once for each client before obtaining the $K$ local updates. Furthermore, since the displacements average to $0$ i.e $\sum_{i=1}^n \vv_i = \sum_{i=1}^n  -\beta\rb{\nabla f(\xx)-\nabla f_i(\xx)} = 0$, they don't require being reverted in the end. This is illustrated through Figure~\ref{fig:FedGA_vs_SCAFFOLD} and further described in the Appendix \ref{app:FedGA}.
We refer to the resulting Algorithm \ref{alg:FedGA}as FedGA (Federated Gradient Alignment).

We assume that, for the $k_{th}$ local update, client $i$ obtains an unbiased stochastic gradient of $f_i$ denoted by $\nabla f_i(.;\zeta_{i,k})$ where $\zeta_{i,k}$ for $k \in [1,\cdots,K]$ are sampled i.i.d  such that $f_i(\xx) := \expect_{\zeta_i}[f_i(\xx;\zeta_i)]$. The stochasticity in the local updates allows our algorithm to retain the generalization benefits of local SGD, while additionally aligning the gradients across clients through the use of suitable displacements.
\begin{algorithm}
  \caption{Federated  Gradient Alignment}%
  \label{alg:FedGA}
  \begin{algorithmic}[1]{
      \State \emph{Input:} Learning rate $\alpha$, initial model parameters $\xx$
      \While{not done}
        \State{$\nabla f(\xx) \gets \frac{1}{n}\sum_{i=1}^{n}\nabla f_i(\xx)$} 
        \Comment{Update the mean gradient computing $\nabla f_i(\xx)$ in parallel} \label{step:update_mean}
        \For{Client $i$ in $[1,\cdots,n]$}
          \State{Obtain the displacement of the mean gradient as $\vv_i \gets \rb{\nabla f(\xx)- \nabla f_i(\xx)}$}
          \State{$\xx^{(0)}_i \gets \xx - \beta\vv_i$}   \Comment{Displacement applied at the beginning}
          \For{$k$ in $[1,\cdots,K]$}
          \State{$\xx^{(k)}_i \gets \xx^{(k-1)}_i -  \alpha\nabla{f_i}(\xx^{(k-1)}_i;\zeta_{i,k})$} 
        
          \EndFor
        \EndFor
      \State{$\xx \gets \frac{1}{n}\sum_{i=1}^{n}\xx^{(K)}_i$}
      \EndWhile
      }
    \end{algorithmic}
\end{algorithm}
Through a derivation similar to Theorem \ref{thm:GA} (Appendix \ref{app:proofs}), we obtain the following result:
\begin{theorem}\label{thm:FedGA regularization}
The expected difference between the parameters reached by FedGA step size $\alpha$ and displacement~$\beta$ and FedAvg  after one round with $K$ local updates per client starting from the initial parameters $\xx$ is given by
\begin{align*}
    &\Ea{\xx_{FedGA} - \xx_{FedAvg}} =\\
    &-\frac{\alpha\beta K}{2n}\nabla_{\xx}\Big(\sum_{i=1}^{n} \norm{\nabla f_i(\xx)-\nabla f(\xx)}^2\Big) +\cO(\alpha\beta^2).
\end{align*}\vspace{-1em}
\end{theorem}
\textbf{Scaffold.}
As noted above, unlike distributed gradient descent with communication at each round, multiple local updates for each client in federated learning can cause the global updates to deviate from the objective's gradient significantly. This motivated \citet{pmlr-v119-karimireddy20a} to use control variate based corrections for each client's local updates. Surprisingly, our analysis reveals that the resulting algorithm, SCAFFOLD, not only minimizes the variance of the updates, but also leads to the alignment of the gradients across clients through implicit regularization. This is because, as illustrated in the Appendix \ref{app:scaffold}, Scaffold and FedGA differ only in that Scaffold directly adds the control variates into the local update while FedGA utilizes them for displacement. This corroborates the empirical improvements in convergence rates and explains the improvements in test accuracies due to SCAFFOLD. 
The implicit gradient alignment in SCAFFOLD is described through the following result, proved in Appendix \ref{app:proofs}:

\begin{theorem}\label{thm:Scaffold regularization}
The expected difference between the parameters reached by SCAFFOLD and FedAvg with step size $\alpha$ after one round with $K$ local updates per client starting from the initial parameters $\xx$ is given by:
\begin{equation}
\begin{split}
    &\xx_{SCAFFOLD} - \xx_{FedAVG} \\&=-\frac{\alpha^2 K(K-1)}{4n}\nabla_{\xx}\Big(\sum_{i=1}^{n} \|\nabla f_i(\xx)-\nabla f(\xx)\|^2\Big) +\cO(\alpha^3).
\end{split}
\end{equation}

\end{theorem}

A crucial difference between FedGA and Scaffold is that FedGA allows the ability to utilize a displacement step size $\beta$, different from $\alpha$, enabling finer control over the effect of the regularization term. Moreover, unlike SCAFFOLD, FedGA does not require applying the displacement at each local step, which improves the consistency between consecutive updates as well as the overall efficiency. We describe this in more detail in Appendix \ref{app:scaffold}.

\begin{table}[h]
	\caption{\small Test Accuracy achieved by FedGA, SCAFFOLD, and FedAvg on EMNIST and CIFAR10. For EMNIST we sample roughly 20\% of the clients in each round, while for CIFAR10 100\% of the clients are used. For EMNIST we distinguish between the IID and the heterogeneous distributions described in Section~\ref{subsec:Exp Federated learning}.\vspace{3mm}}
	\centering
	\resizebox{0.5\textwidth}{!}{
	\begin{tabular}{lccc}
	\toprule
	\parbox{2cm}{}   			& \parbox{2cm}{\centering EMNIST\\ IID \\ 10 out of 47}		&\!\!\!\parbox{2.5cm}{\centering EMNIST\\ heterogeneous \\ 10 out of 47}\!\!&\!\! \parbox{2.5cm}{\centering CIFAR10 \\ IID \\ 10 out of 10} \\ \midrule
		FedGA		& \bm{$88.66 \pm 0.13$}		& $\bm{85.95 \pm 0.56}$			& $\bm{74.34 \pm 0.48}$	\\ \midrule
		SCAFFOLD	& $88.56 \pm 0.12$		& $84.67 \pm 0.78$			& $73.89 \pm 0.65$	\\ \midrule
		FedAvg		& $88.32 \pm 0.06$		& $82.9 \pm 0.58$			& $73.1 \pm 0.17$ \\ \midrule
		FedProx & $88.176 \pm 0.12$ & $83.197 \pm 0.19$  & $73.93 \pm 0.38$\\ \midrule
    \bottomrule
    \label{table:Main Results}
	\end{tabular}
	}
\end{table}

\vspace{-.1in}
\section{Experiments}\label{sec:Experiments}

Motivated by the analysis presented in previous sections, we aim to confirm the effectiveness of implicit regularization through a series of experiments on image classification tasks. 
To this end, we evaluate the effectiveness of GradAlign in achieving improved generalization in the following settings:
$(1)$ Federated Learning: Data is distributed on a large number of clients (with different distributions), and only a subset of the clients is sampled to be used in each round. $(2)$ Datacenter distributed learning: Data is distributed (i.i.d.) among the clients, and all clients are used on each round.

Since our primary focus is the quantitative evaluation of generalization performance through test accuracy and test losses, we do not constrain the algorithms to use the same number of local epochs (a local epoch is completed when the entire data of a client has been used, typically in Federated Learning a client can pass more than once trough its data before communicating). 
Indeed, while increasing the number of local epochs may decrease the number of rounds needed to train, it has no noticeable effect on the maximum test accuracy reached by the algorithm (see Appendix~\ref{app:using_more_local_epochs}).
To further verify the regularization effects of our approach, we provide comparisons of training accuracies in Appendix \ref{app:training accuracy}, showing that the improvements due to gradient alignment are largely in the test rather than training loss. 
We use a constant learning rate throughout all our experiments to illustrate, as has been done in several federated learning papers~\citep{mcmahan2017communication,hsu2019measuring,khaled2020tighter,liu2020accelerating}. 
We also do not use batch normalization or momentum (neither server nor local momentum) in our experiments.
Throughout, we report the best results with the hyperparameters obtained through grid search for each of the studied algorithms. For more details, see Appendix~\ref{app:hyperparameters}.
Moreover, each of the reported curves and results is averaged over at least 3 different runs with different random seeds. 

All experiments were performed using PyTorch on Tesla V100-SXM2 with 32GB of memory. 

Recall that both FedGA and SCAFFOLD require one extra round of communication to compute the displacement. \textbf{This extra round is included in all our plots and results}, i.e., even with this $2\times$ overhead, FedGA still outperforms the competition.
To ensure a fair comparison, for both the settings, we use the following definition of rounds:

\paragraph{Definition of Rounds}: In our experimental plots (Figures \ref{fig:cifar_dist}, \ref{fig:gradient_variance}, \ref{fig:emnist_iid}), the "rounds" label in the x-axis denotes the total communication rounds, including the extra communication round for computing the displacements. Thus while the communication cost can be further reduced by utilizing estimates of the mean gradient, similar to "Option II" in SCAFFOLD \citep{pmlr-v119-karimireddy20a}, our results clearly demonstrate that even without such approximations, FedGA can be used to improve generalization in practical federated learning settings without a significant communication overhead.

\subsection{Federated Learning}\label{subsec:Exp Federated learning}
For Federated learning, we use the (balanced) EMNIST dataset~\citep{cohen2017emnist} consisting of 47 classes distributed among 47 clients, each receiving 2400 training examples. 
We split the data using two distinct distributions: In the \emph{IID} setting, data is shuffled using a random permutation and then distributed (without overlap) among the 47 clients.
In the \emph{heterogeneous} setting, each of the 47 clients is assigned all the data corresponding to a unique label from the 47 classes. This setting has been extensively studied following the work of~\citet{hsu2019measuring}.
We further include additional results on Natural Language Processing tasks in Appendix \ref{app:NLP results} and CIFAR-100 in Appendix~ \ref{app:CIFAR100} along with plots of the variance of gradients and test accuracy for EMNIST in Appendix~\ref{app:plots}.
For EMNIST, we use a (simple) CNN neural network architecture for our experiments with 2 convolutional layers followed by a fully connected layer. 
The exact description of the network can be found in the Appendix~\ref{app:Architectures}.
In each round, we sample 10 out of 47 clients uniformly at random. We compare the performance of four algorithms: FedAvg, Scaffold, FedProx \citep{fedprox}, and FedGA.
With approximately 20\% of the clients sampled on each round, FedGA achieves the highest Test accuracy and the lowest Test Loss in both settings (see Figure~\ref{fig:emnist_iid}).

\paragraph{IID data} 
Since the data in each client is i.i.d. sampled, using smaller mini-batches for local steps achieves an implicit regularization that promotes gradient alignment within the clients' data (see Section~\ref{sec:k sequential steps}). Scaffold, FedGA, and FedAvg all benefit from this regularization when using smaller mini-batches. 
On top of that, FedGA and Scaffold promote inter-client gradient alignment as seen in Theorems~\ref{thm:FedGA regularization} and~\ref{thm:Scaffold regularization}. 
Therefore, these algorithms with smaller mini-batches benefit from both inter and intra client gradient alignment. We believe this is the reason why they clearly outperform FedAvg; see Figure~\ref{fig:emnist_iid}.
Furthermore, FedGA has an additional parameter $\beta$ that can be used to tune the constant in front of the regularizer (see Theorem~\ref{thm:FedGA regularization}). Thus, while the implicit regularization term might be present in both Scaffold and FedGA, the fine-tuning of this parameter is crucial for its improvements over Scaffold. 
Indeed, as seen in Appendix~\ref{app:impact on tuning beta}, modifying the constant $\beta$ has a significant impact on the performance of FedGA.
This is a double-edged sword, where on the one hand, $\beta$ improves generalization, but on the other hand, it can be quite difficult to tune. 
In fact, $\beta$ used for the IID and the heterogeneous settings are different, as they depend on the magnitude of the displacement.

\paragraph{Heterogeneous data} 
Federated learning is more challenging if each client has their own data distribution~\citep{hsu2019measuring}, as the gradients become less transferable between clients. Achieving gradient alignment thus has a strong promise to mitigate this problem and to better align the updates on clients with the common objective. 
Indeed, FedGA achieves a significantly better generalization than FedAvg and SCAFFOLD, the latter ranking in the middle but closer to FedAvg. 
We also found that increasing the batch size had only a minor impact on training with FedAvg, while it significantly impacts FedGA and SCAFFOLD. 

\subsection{Datacenter distributed learning}
We use the CIFAR10 dataset~\citep{krizhevsky2009learning} consisting of 50000 training examples split among 10 classes, which are then distributed among 10 clients, each receiving 5000 training examples.
We split the data using the same IID setting used in Federated Learning.
We use a (simple) CNN neural network architecture consisting of 2 convolutional layers followed by 2 fully connected layers. The exact description of the network can be found in the supplementary materials.
We study two different settings: In the first, we are interested in maximizing parallelism, i.e., we assume that communication is not the bottleneck, and hence we aim to minimize the total number of updates to reach top accuracy, while communicating once per local gradient computation. In this setting we compare GradAlign (FedGA with $K=1$) against large-batch SGD and SCAFFOLD (large-batch).
The second setting is equivalent to the IID federated learning setting, but with every client sampled in each round.

\begin{figure}[ht]
\centering
\includegraphics[width=0.49\textwidth]{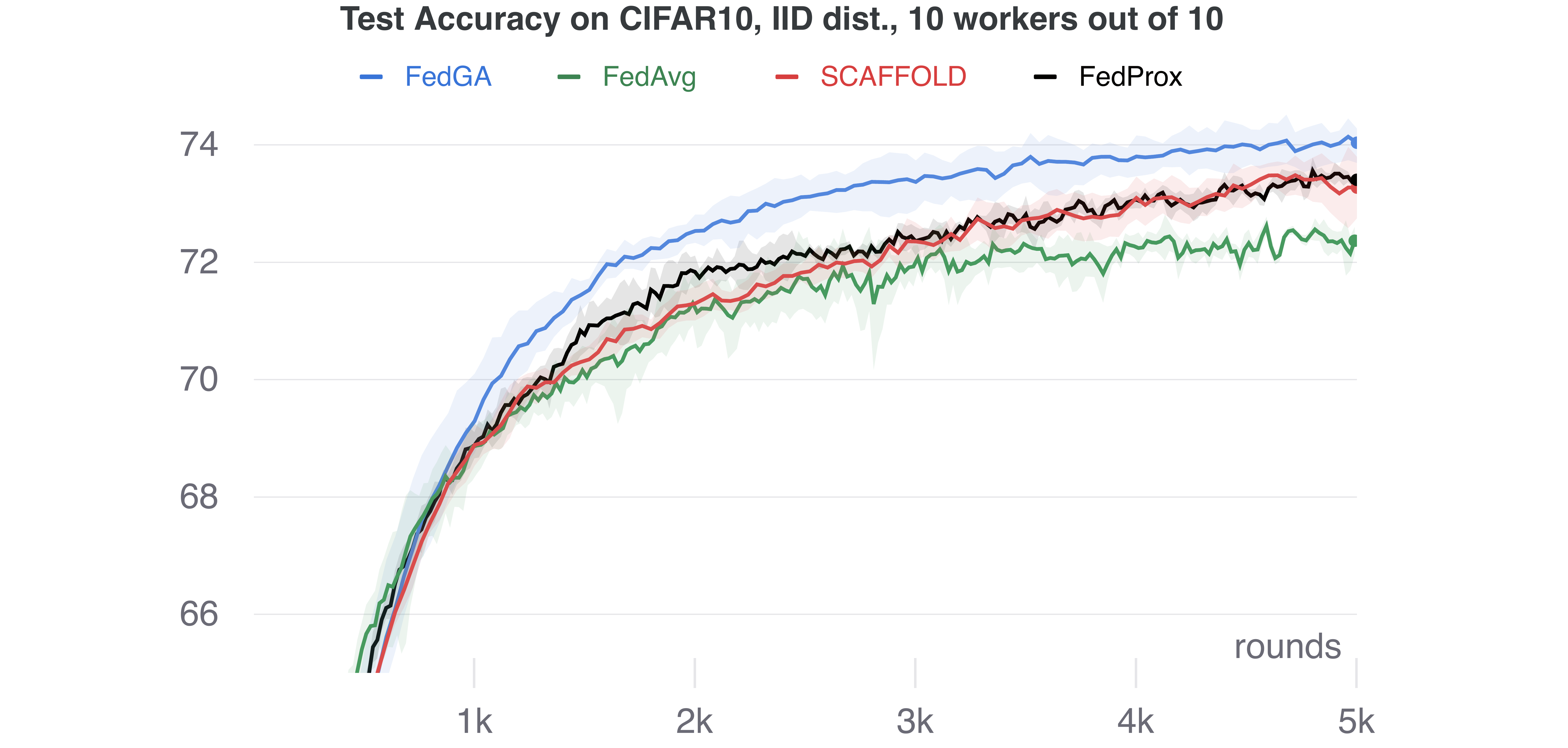}
\includegraphics[width=0.49\textwidth]{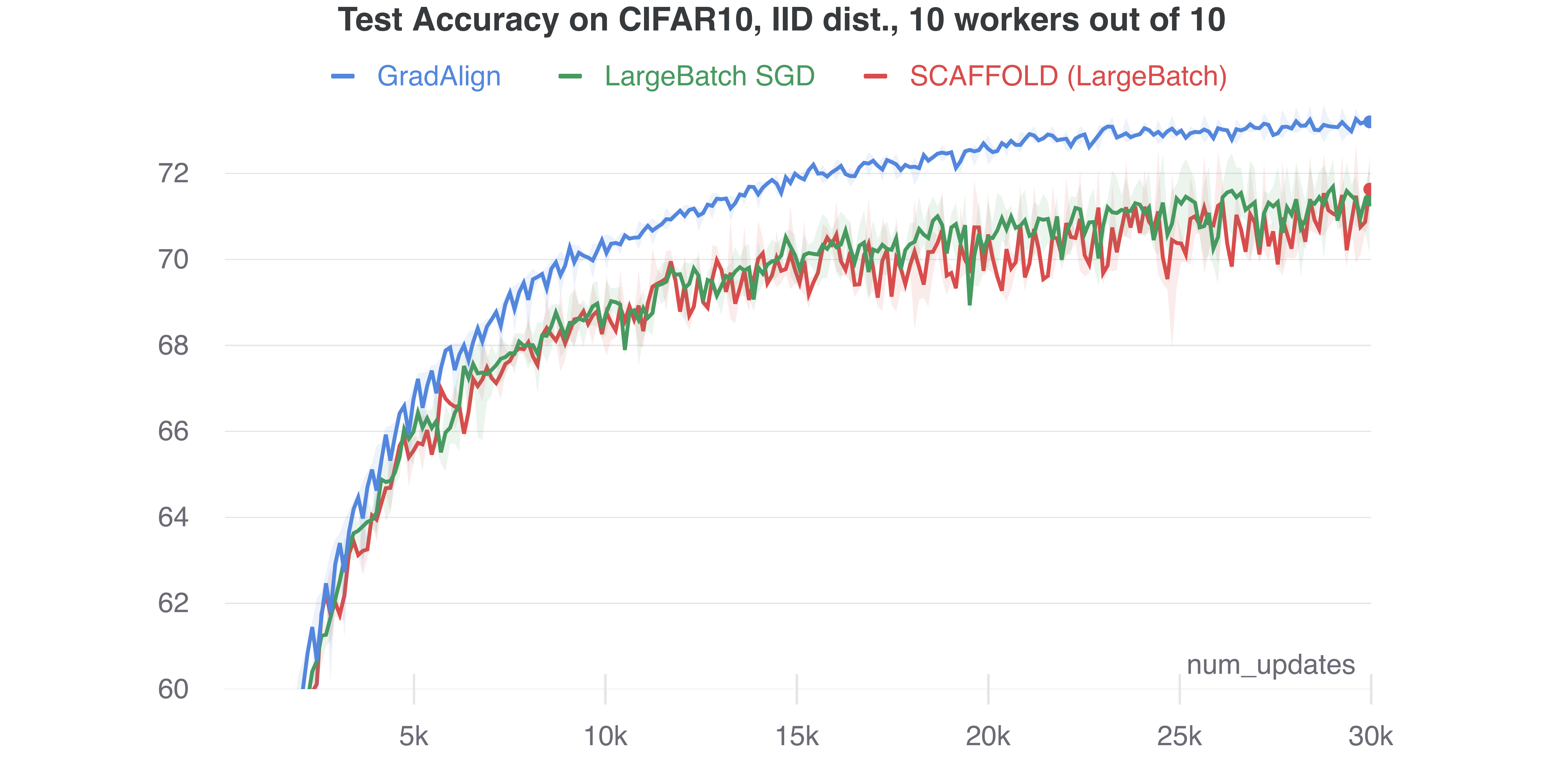}\vspace{-2mm}
\caption{\small Test accuracy on the CIFAR10 dataset using a CNN for the distributed setting where 100\% client sampling per round.
Top: In the Federated Learning setting FedGA is not only faster in terms of the number of rounds, but it also achieves higher test accuracy than its counterparts.
Bottom: The $x$-axis depicts the number of updates, i.e., the number of times the parameters of the model are modified. With this metric, GradAlign profits from the available parallelism better than Large-Batch SGD and SCAFFOLD.}
\vspace{-4mm}
\label{fig:cifar_dist}
\end{figure}

\paragraph{Sampling all clients}
Similar to the IID federated learning setting, FedGA obtains the highest accuracy followed by SCAFFOLD and then by FedAvg; see Figure~\ref{fig:cifar_dist}. 
In this setting, even with the overhead of $2\times$ in the number of rounds used by both FedGA and Scaffold, they outperform FedAvg.
As in the federated IID setting, a smaller mini-batch size benefits all algorithms. 
We believe this is explained by the gradient alignment coming from the use of different mini-batches sequentially during the local updates.
In this way, both FedGA and SCAFFOLD benefit from inter- and intra-client gradient alignment.

\paragraph{Minimizing number of updates.} 
In this setting, the algorithm to beat is Large-Batch SGD. If communication is fast enough, the main bottleneck is the sequential dependencies between consecutive gradient updates. 
To increase parallelism, the standard solution is to increase the batch size, but it is known to have an impact on generalization~\citep{DBLP:conf/iclr/KeskarMNST17,pmlr-v80-ma18a,pmlr-v84-yin18a}. 
Our algorithm GradAlign (see Section~\ref{section:GradAlign}) allows us to use large mini-batches while retaining the generalization properties of using smaller mini-batches.
Indeed, our experiments show that GradAlign noticeably achieves higher Test Accuracy than Large-Batch SGD. Moreover, it converges faster in terms of the number of updates (see Figure~\ref{fig:cifar_dist}).

\vspace{-2mm}

\vspace{-1mm}
\section{Future Work}\label{sec:future_limit}
Promising directions for future work include designing algorithms with implicit gradient alignment for decentralized and asynchronous learning settings, incorporating optimization schemes such as momentum into gradient alignment, and developing techniques to reduce the communication overhead in FedGA. 

\newpage
\begin{small}
\bibliography{aaai}

\begin{thebibliography}{43}
\providecommand{\natexlab}[1]{#1}

\bibitem[{Acar et~al.(2021)Acar, Zhao, Matas, Mattina, Whatmough, and
  Saligrama}]{feddyn}
Acar, D. A.~E.; Zhao, Y.; Matas, R.; Mattina, M.; Whatmough, P.; and Saligrama,
  V. 2021.
\newblock Federated Learning Based on Dynamic Regularization.
\newblock In \emph{International Conference on Learning Representations}.

\bibitem[{Barrett and Dherin(2021)}]{barrett2021implicit}
Barrett, D.; and Dherin, B. 2021.
\newblock Implicit Gradient Regularization.
\newblock In \emph{International Conference on Learning Representations}.

\bibitem[{Caldas et~al.(2018)Caldas, Duddu, Wu, Li, Kone{\v{c}}n{\`y}, McMahan,
  Smith, and Talwalkar}]{caldas2018leaf}
Caldas, S.; Duddu, S. M.~K.; Wu, P.; Li, T.; Kone{\v{c}}n{\`y}, J.; McMahan,
  H.~B.; Smith, V.; and Talwalkar, A. 2018.
\newblock Leaf: A benchmark for federated settings.
\newblock \emph{arXiv preprint arXiv:1812.01097}.

\bibitem[{Chatterjee(2020)}]{Chatterjee2020Coherent}
Chatterjee, S. 2020.
\newblock Coherent Gradients: An Approach to Understanding Generalization in
  Gradient Descent-based Optimization.
\newblock In \emph{International Conference on Learning Representations}.

\bibitem[{Chatterjee and Zielinski(2020)}]{chatterjee2020making}
Chatterjee, S.; and Zielinski, P. 2020.
\newblock Making Coherence Out of Nothing At All: Measuring the Evolution of
  Gradient Alignment.
\newblock arXiv:2008.01217.

\bibitem[{Chaudhari and Soatto(2018)}]{chaudhari2018stochastic}
Chaudhari, P.; and Soatto, S. 2018.
\newblock Stochastic gradient descent performs variational inference, converges
  to limit cycles for deep networks.
\newblock In \emph{International Conference on Learning Representations}.

\bibitem[{Cohen et~al.(2017)Cohen, Afshar, Tapson, and
  Van~Schaik}]{cohen2017emnist}
Cohen, G.; Afshar, S.; Tapson, J.; and Van~Schaik, A. 2017.
\newblock EMNIST: Extending MNIST to handwritten letters.
\newblock In \emph{2017 International Joint Conference on Neural Networks
  (IJCNN)}, 2921--2926. IEEE.

\bibitem[{Dean et~al.(2012)Dean, Corrado, Monga, Chen, Devin, Mao, Ranzato,
  Senior, Tucker, Yang, Le, and Ng}]{NIPS2012_6aca9700}
Dean, J.; Corrado, G.; Monga, R.; Chen, K.; Devin, M.; Mao, M.; Ranzato, M.~a.;
  Senior, A.; Tucker, P.; Yang, K.; Le, Q.; and Ng, A. 2012.
\newblock Large Scale Distributed Deep Networks.
\newblock In Pereira, F.; Burges, C. J.~C.; Bottou, L.; and Weinberger, K.~Q.,
  eds., \emph{Advances in Neural Information Processing Systems}, volume~25.
  Curran Associates, Inc.

\bibitem[{Dinh et~al.(2017)Dinh, Pascanu, Bengio, and
  Bengio}]{pmlr-v70-dinh17b}
Dinh, L.; Pascanu, R.; Bengio, S.; and Bengio, Y. 2017.
\newblock Sharp Minima Can Generalize For Deep Nets.
\newblock In Precup, D.; and Teh, Y.~W., eds., \emph{Proceedings of the 34th
  International Conference on Machine Learning}, volume~70 of \emph{Proceedings
  of Machine Learning Research}, 1019--1028. PMLR.

\bibitem[{Fort et~al.(2020)Fort, Nowak, Jastrzebski, and
  Narayanan}]{fort2020stiffness}
Fort, S.; Nowak, P.~K.; Jastrzebski, S.; and Narayanan, S. 2020.
\newblock Stiffness: A New Perspective on Generalization in Neural Networks.
\newblock arXiv:1901.09491.

\bibitem[{Goyal et~al.(2018)Goyal, Dollár, Girshick, Noordhuis, Wesolowski,
  Kyrola, Tulloch, Jia, and He}]{goyal2018accurate}
Goyal, P.; Dollár, P.; Girshick, R.; Noordhuis, P.; Wesolowski, L.; Kyrola,
  A.; Tulloch, A.; Jia, Y.; and He, K. 2018.
\newblock Accurate, Large Minibatch SGD: Training ImageNet in 1 Hour.
\newblock arXiv:1706.02677.

\bibitem[{Hsu, Qi, and Brown(2019{\natexlab{a}})}]{fedavgm}
Hsu, T.-M.~H.; Qi, H.; and Brown, M. 2019{\natexlab{a}}.
\newblock Measuring the Effects of Non-Identical Data Distribution for
  Federated Visual Classification.
\newblock arXiv:1909.06335.

\bibitem[{Hsu, Qi, and Brown(2019{\natexlab{b}})}]{hsu2019measuring}
Hsu, T.-M.~H.; Qi, H.; and Brown, M. 2019{\natexlab{b}}.
\newblock Measuring the effects of non-identical data distribution for
  federated visual classification.
\newblock \emph{arXiv preprint arXiv:1909.06335}.

\bibitem[{Jacot, Gabriel, and Hongler(2018)}]{NEURIPS2018_5a4be1fa}
Jacot, A.; Gabriel, F.; and Hongler, C. 2018.
\newblock Neural Tangent Kernel: Convergence and Generalization in Neural
  Networks.
\newblock In Bengio, S.; Wallach, H.; Larochelle, H.; Grauman, K.;
  Cesa-Bianchi, N.; and Garnett, R., eds., \emph{Advances in Neural Information
  Processing Systems}, volume~31. Curran Associates, Inc.

\bibitem[{Jastrzębski et~al.(2018)Jastrzębski, Kenton, Arpit, Ballas,
  Fischer, Bengio, and Storkey}]{jastrzebski2018factors}
Jastrzębski, S.; Kenton, Z.; Arpit, D.; Ballas, N.; Fischer, A.; Bengio, Y.;
  and Storkey, A. 2018.
\newblock Three Factors Influencing Minima in SGD.
\newblock arXiv:1711.04623.

\bibitem[{Johnson and Zhang(2013)}]{DBLP:conf/nips/Johnson013}
Johnson, R.; and Zhang, T. 2013.
\newblock Accelerating Stochastic Gradient Descent using Predictive Variance
  Reduction.
\newblock In \emph{NIPS}, 315--323.

\bibitem[{Kairouz and McMahan(2021)}]{MAL-083}
Kairouz, P.; and McMahan, H.~B. 2021.
\newblock Advances and Open Problems in Federated Learning.
\newblock \emph{Foundations and Trends® in Machine Learning}, 14(1): --.

\bibitem[{Kairouz et~al.(2019)Kairouz, McMahan, Avent, Bellet, Bennis, Bhagoji,
  Bonawitz, Charles, Cormode, Cummings et~al.}]{kairouz2019advances}
Kairouz, P.; McMahan, H.~B.; Avent, B.; Bellet, A.; Bennis, M.; Bhagoji, A.~N.;
  Bonawitz, K.; Charles, Z.; Cormode, G.; Cummings, R.; et~al. 2019.
\newblock Advances and open problems in federated learning.
\newblock \emph{arXiv preprint arXiv:1912.04977}.

\bibitem[{Karimireddy et~al.(2020)Karimireddy, Kale, Mohri, Reddi, Stich, and
  Suresh}]{pmlr-v119-karimireddy20a}
Karimireddy, S.~P.; Kale, S.; Mohri, M.; Reddi, S.; Stich, S.; and Suresh,
  A.~T. 2020.
\newblock {SCAFFOLD}: Stochastic Controlled Averaging for Federated Learning.
\newblock In III, H.~D.; and Singh, A., eds., \emph{Proceedings of the 37th
  International Conference on Machine Learning}, volume 119 of
  \emph{Proceedings of Machine Learning Research}, 5132--5143. PMLR.

\bibitem[{Keskar et~al.(2017)Keskar, Mudigere, Nocedal, Smelyanskiy, and
  Tang}]{DBLP:conf/iclr/KeskarMNST17}
Keskar, N.~S.; Mudigere, D.; Nocedal, J.; Smelyanskiy, M.; and Tang, P. T.~P.
  2017.
\newblock On Large-Batch Training for Deep Learning: Generalization Gap and
  Sharp Minima.
\newblock In \emph{5th International Conference on Learning Representations,
  {ICLR} 2017, Toulon, France, April 24-26, 2017, Conference Track
  Proceedings}. OpenReview.net.

\bibitem[{Khaled, Mishchenko, and Richt{\'a}rik(2020)}]{khaled2020tighter}
Khaled, A.; Mishchenko, K.; and Richt{\'a}rik, P. 2020.
\newblock Tighter theory for local SGD on identical and heterogeneous data.
\newblock In \emph{International Conference on Artificial Intelligence and
  Statistics}, 4519--4529. PMLR.

\bibitem[{Konečný et~al.(2016)Konečný, McMahan, Ramage, and
  Richtárik}]{2016federated}
Konečný, J.; McMahan, H.~B.; Ramage, D.; and Richtárik, P. 2016.
\newblock Federated Optimization: Distributed Machine Learning for On-Device
  Intelligence.
\newblock arXiv:1610.02527.

\bibitem[{Krizhevsky, Hinton et~al.(2009)}]{krizhevsky2009learning}
Krizhevsky, A.; Hinton, G.; et~al. 2009.
\newblock Learning multiple layers of features from tiny images.

\bibitem[{Li et~al.(2020)Li, Sahu, Zaheer, Sanjabi, Talwalkar, and
  Smith}]{fedprox}
Li, T.; Sahu, A.~K.; Zaheer, M.; Sanjabi, M.; Talwalkar, A.; and Smith, V.
  2020.
\newblock Federated Optimization in Heterogeneous Networks.
\newblock In Dhillon, I.~S.; Papailiopoulos, D.~S.; and Sze, V., eds.,
  \emph{Proceedings of Machine Learning and Systems 2020, MLSys 2020, Austin,
  TX, USA, March 2-4, 2020}. mlsys.org.

\bibitem[{Lin et~al.(2020{\natexlab{a}})Lin, Kong, Stich, and
  Jaggi}]{lin2020extrapolation}
Lin, T.; Kong, L.; Stich, S.; and Jaggi, M. 2020{\natexlab{a}}.
\newblock Extrapolation for Large-batch Training in Deep Learning.
\newblock In \emph{ICML - Proceedings of the 37th International Conference on
  Machine Learning}, volume 119 of \emph{Proceedings of Machine Learning
  Research}, 6094--6104. PMLR.

\bibitem[{Lin et~al.(2020{\natexlab{b}})Lin, Stich, Patel, and
  Jaggi}]{Lin2020Don't}
Lin, T.; Stich, S.~U.; Patel, K.~K.; and Jaggi, M. 2020{\natexlab{b}}.
\newblock Don't Use Large Mini-batches, Use Local SGD.
\newblock In \emph{International Conference on Learning Representations}.

\bibitem[{Liu et~al.(2020)Liu, Chen, Chen, and Zhang}]{liu2020accelerating}
Liu, W.; Chen, L.; Chen, Y.; and Zhang, W. 2020.
\newblock Accelerating federated learning via momentum gradient descent.
\newblock \emph{IEEE Transactions on Parallel and Distributed Systems}, 31(8):
  1754--1766.

\bibitem[{Ma, Bassily, and Belkin(2018)}]{pmlr-v80-ma18a}
Ma, S.; Bassily, R.; and Belkin, M. 2018.
\newblock The Power of Interpolation: Understanding the Effectiveness of {SGD}
  in Modern Over-parametrized Learning.
\newblock In Dy, J.; and Krause, A., eds., \emph{Proceedings of the 35th
  International Conference on Machine Learning}, volume~80 of \emph{Proceedings
  of Machine Learning Research}, 3325--3334. PMLR.

\bibitem[{Mandt, Hoffman, and Blei(2017)}]{mandtsgd}
Mandt, S.; Hoffman, M.~D.; and Blei, D.~M. 2017.
\newblock Stochastic Gradient Descent as Approximate Bayesian Inference.
\newblock 18(1): 4873–4907.

\bibitem[{McMahan et~al.(2017{\natexlab{a}})McMahan, Moore, Ramage, Hampson,
  and y~Arcas}]{pmlr-v54-mcmahan17a}
McMahan, B.; Moore, E.; Ramage, D.; Hampson, S.; and y~Arcas, B.~A.
  2017{\natexlab{a}}.
\newblock {Communication-Efficient Learning of Deep Networks from Decentralized
  Data}.
\newblock In Singh, A.; and Zhu, J., eds., \emph{Proceedings of the 20th
  International Conference on Artificial Intelligence and Statistics},
  volume~54 of \emph{Proceedings of Machine Learning Research}, 1273--1282.
  Fort Lauderdale, FL, USA: PMLR.

\bibitem[{McMahan et~al.(2017{\natexlab{b}})McMahan, Moore, Ramage, Hampson,
  and y~Arcas}]{mcmahan2017communication}
McMahan, B.; Moore, E.; Ramage, D.; Hampson, S.; and y~Arcas, B.~A.
  2017{\natexlab{b}}.
\newblock Communication-Efficient Learning of Deep Networks from Decentralized
  Data.
\newblock In \emph{Proceedings of AISTATS}, 1273--1282.

\bibitem[{Nedic(2020)}]{nedic2020review}
Nedic, A. 2020.
\newblock Distributed Gradient Methods for Convex Machine Learning Problems in
  Networks: Distributed Optimization.
\newblock \emph{{IEEE} Signal Processing Magazine}, 37(3): 92--101.

\bibitem[{Nichol, Achiam, and Schulman(2018)}]{nichol2018firstorder}
Nichol, A.; Achiam, J.; and Schulman, J. 2018.
\newblock On First-Order Meta-Learning Algorithms.
\newblock arXiv:1803.02999.

\bibitem[{Reddi et~al.(2021)Reddi, Charles, Zaheer, Garrett, Rush,
  Kone{\v{c}}n{\'y}, Kumar, and McMahan}]{reddi2021adaptive}
Reddi, S.~J.; Charles, Z.; Zaheer, M.; Garrett, Z.; Rush, K.;
  Kone{\v{c}}n{\'y}, J.; Kumar, S.; and McMahan, H.~B. 2021.
\newblock Adaptive Federated Optimization.
\newblock In \emph{International Conference on Learning Representations}.

\bibitem[{Robbins and Monro(1951)}]{robbins1951sgd}
Robbins, H.; and Monro, S. 1951.
\newblock {A Stochastic Approximation Method}.
\newblock \emph{The Annals of Mathematical Statistics}, 22(3): 400--407.

\bibitem[{Schmidt and Roux(2013)}]{schmidt2013fast}
Schmidt, M.; and Roux, N.~L. 2013.
\newblock Fast Convergence of Stochastic Gradient Descent under a Strong Growth
  Condition.
\newblock arXiv:1308.6370.

\bibitem[{Shallue et~al.(2019)Shallue, Lee, Antognini, Sohl-Dickstein, Frostig,
  and Dahl}]{shallue2019batch}
Shallue, C.~J.; Lee, J.; Antognini, J.; Sohl-Dickstein, J.; Frostig, R.; and
  Dahl, G.~E. 2019.
\newblock Measuring the Effects of Data Parallelism on Neural Network Training.
\newblock \emph{Journal of Machine Learning Research}, 20(112): 1--49.

\bibitem[{Smith et~al.(2021)Smith, Dherin, Barrett, and De}]{smith2021on}
Smith, S.~L.; Dherin, B.; Barrett, D.; and De, S. 2021.
\newblock On the Origin of Implicit Regularization in Stochastic Gradient
  Descent.
\newblock In \emph{International Conference on Learning Representations}.

\bibitem[{Smith and Le(2018)}]{l.2018a}
Smith, S.~L.; and Le, Q.~V. 2018.
\newblock A Bayesian Perspective on Generalization and Stochastic Gradient
  Descent.
\newblock In \emph{International Conference on Learning Representations}.

\bibitem[{Woodworth et~al.(2020)Woodworth, Patel, Stich, Dai, Bullins, Mcmahan,
  Shamir, and Srebro}]{pmlr-v119-woodworth20a}
Woodworth, B.; Patel, K.~K.; Stich, S.; Dai, Z.; Bullins, B.; Mcmahan, B.;
  Shamir, O.; and Srebro, N. 2020.
\newblock Is Local {SGD} Better than Minibatch {SGD}?
\newblock In III, H.~D.; and Singh, A., eds., \emph{Proceedings of the 37th
  International Conference on Machine Learning}, volume 119 of
  \emph{Proceedings of Machine Learning Research}, 10334--10343. PMLR.

\bibitem[{Yao et~al.(2018)Yao, Gholami, Keutzer, and Mahoney}]{YaoHessian}
Yao, Z.; Gholami, A.; Keutzer, K.; and Mahoney, M.~W. 2018.
\newblock Hessian-Based Analysis of Large Batch Training and Robustness to
  Adversaries.
\newblock In \emph{Proceedings of the 32nd International Conference on Neural
  Information Processing Systems}, NIPS'18, 4954–4964. Red Hook, NY, USA:
  Curran Associates Inc.

\bibitem[{Yin et~al.(2018)Yin, Pananjady, Lam, Papailiopoulos, Ramchandran, and
  Bartlett}]{pmlr-v84-yin18a}
Yin, D.; Pananjady, A.; Lam, M.; Papailiopoulos, D.; Ramchandran, K.; and
  Bartlett, P. 2018.
\newblock Gradient Diversity: a Key Ingredient for Scalable Distributed
  Learning.
\newblock In Storkey, A.; and Perez-Cruz, F., eds., \emph{Proceedings of the
  Twenty-First International Conference on Artificial Intelligence and
  Statistics}, volume~84 of \emph{Proceedings of Machine Learning Research},
  1998--2007. PMLR.

\bibitem[{Zinkevich et~al.(2010)Zinkevich, Weimer, Smola, and Li}]{parallelSGD}
Zinkevich, M.~A.; Weimer, M.; Smola, A.; and Li, L. 2010.
\newblock Parallelized Stochastic Gradient Descent.
\newblock NIPS'10, 2595–2603. Red Hook, NY, USA: Curran Associates Inc.

\end{thebibliography}
\end{small}
\newpage
\appendix
\onecolumn
\section{Appendix}

\subsection{Descent condition}\label{app:descent condition}

In this section, we provide sufficient conditions for the smoothness of the regularizer $r(\xx)$ and subsequently prove Theorem \ref{thm:descent}.

\subsubsection{Smoothness of Variance}
While the smoothness of the objective $f(\xx)$ is commonly used to prove the sufficient conditions for descent (decrease of the objective value) in general non-convex settings, the smoothness of the variance regularization term $r(\xx)$ requires a few additional assumptions as 
illustrated through the subsequent analysis.
The term $\norm{\nabla r(\xx)- \nabla r(\yy)}$ can be bounded as follows:
      \begin{align*}
          &\norm{\nabla r(\xx)- \nabla r(\yy)}\\
          &= \norm{\frac{1}{n}\sum_{i=1}^{n}(\nabla^2 f_i(\xx)-\nabla^2 f(\xx))(\nabla f_i(\xx)-\nabla f(\xx)))-\frac{1}{n}\sum_{i=1}^{n}(\nabla^2 f_i(\yy)-\nabla^2 f(\yy))(\nabla f_i(\yy)-\nabla f(\yy)))}\\
          &\leq \norm{\frac{1}{n}\sum_{i=1}^{n}(\nabla^2 f_i(\xx)-\nabla^2 f(\xx))\rb{\rb{\nabla f_i(\xx)-\nabla f(\xx)}-\rb{\nabla f_i(\yy)-\nabla f(\yy)})}}
          \\&+\norm{\frac{1}{n}\sum_{i=1}^{n}\rb{\rb{\nabla^2 f_i(\xx)-\nabla^2 f(\xx)}-\rb{\nabla^2 f_i(\yy)-\nabla^2 f(\yy)}}(\nabla f_i(\yy)-\nabla f(\yy)))}.
      \end{align*}
Thus boundedness and Lipschitzness of $\nabla f_i(\xx)-\nabla f(\xx)$ and, $\nabla^2 f_i(\xx)$ are sufficient conditions for the smoothness of $r(\xx)$. 
Moreover, since the positivity of $\norm{\nabla f_i(\xx)-\nabla f(\xx)}$ and the Cauchy–Schwarz inequality further imply that
\begin{equation*}
    \norm{\nabla f_i(\xx)-\nabla f(\xx)} \leq \sum_{j=1}^n \norm{\nabla f_j(\xx)-\nabla f(\xx)} \leq \sqrt{n}\rb{\sum_{j=1}^n \norm{\nabla f_j(\xx)-\nabla f(\xx)}^2}^{\frac{1}{2}},
\end{equation*}
we note that boundedness of $\norm{\nabla f_i(\xx)-\nabla f(\xx)}$ also follows from the boundedness of variance.

\subsubsection{Theorem \ref{thm:descent}}
\begin{proof}
 Using the $L_1, L_2$ smoothness of $f(\xx),r(\xx)$ respectively and $\nabla \hat{f}(\xx) = \nabla f(\xx) + \beta \nabla r(\xx)$, we have:
 \begin{equation}
 \label{eq:smoothFedGA}
    \hat{f}(\xt{t+1}) - \hat{f}(\xt{t}) \leq \lin*{\xt{t+1}-\xt{t}}{\nabla \hat{f}(\xt{t})} + \frac{L_1}{2}\norm{\xt{t+1}-\xt{t}}^2 + \frac{\beta L_2}{2}\norm{\xt{t+1}-\xt{t}}^2,
 \end{equation}
where $\lin*{\cdot}{\cdot}$ denotes the standard inner product in Euclidean space.
Following the notation in Section \ref{sec:main}, we denote by $\vv_i$, the displacement $-\beta\rbr*{\nabla f(\xt{t})-\nabla f_i(\xt{t})}$ corresponding to the $i_{th}$ minibatch. Using the fundamental theorem of calculus applied to each component of $\nabla f_i$, we can express $\xt{t+1} - \xt{t}$ as follows:
\begin{equation}
\label{eq:taylorerror}
  \begin{gathered}[b]
   \xt{t+1} - \xt{t} = - \alpha \rb{\frac{1}{n}\sum_{i=1}^n \nabla f_i(\xt{t}+\vv_i)}\\
   = - \alpha \frac{1}{n}\sum_{i=1}^n \rb{\nabla f_i(\xt{t})+\nabla^2f_i(\xt{t})\rb{\vv_i} + \int_{z=0}^1 \rb{\nabla^2f_i(\xt{t}+z\vv_i) - \nabla^2f_i(\xt{t})}\vv_idz }\\
   = -\alpha\nabla \hat{f}(\xt{t}) - \alpha \frac{1}{n}\sum_{i=1}^n \int_{z=0}^1 \rb{\nabla^2f_i(\xt{t}+z\vv_i) - \nabla^2f_i(\xt{t})}\vv_idz.
  \end{gathered}
\end{equation}

We now utilize the above expression to bound the terms in equation \eqref{eq:smoothFedGA} as follows:
\begin{align*}
     &\lin*{\rb{\xt{t+1}-\xt{t}}}{\nabla \hat{f}(\xt{t})} = \frac{1}{\alpha}\lin*{\xt{t+1}-\xt{t}}{-\rb{\xt{t+1}-\xt{t}}+\alpha\nabla \hat{f}(\xt{t})+\rb{\xt{t+1}-\xt{t}}}\\
     &=-\frac{1}{\alpha}\norm{\xt{t+1}-\xt{t}}^2-\frac{1}{n}\sum_{i=1}^n\lin*{\int_{z=0}^1 \rb{\nabla^2f_i(\xt{t}+z\vv_i) - \nabla^2f_i(\xt{t})}\vv_idz}{\rbr*{\xt{t+1}-\xt{t}}}\\
     &= -\frac{1}{\alpha}\norm{\xt{t+1}-\xt{t}}^2-\frac{1}{n}\sum_{i=1}^n\lin*{\int_{z=0}^\beta \rb{\nabla^2f_i(\xt{t}+z\vv_i) - \nabla^2f_i(\xt{t})}\vv_idz}{\rbr*{\xt{t+1}-\xt{t}}}\\
     & =-\frac{1}{\alpha}\norm{\xt{t+1}-\xt{t}}^2-\frac{1}{n}\sum_{i=1}^n\int_{z=0}^1\lin*{ \rb{\nabla^2f_i(\xt{t}+z\vv_i) - \nabla^2f_i(\xt{t})}\uu_i}{\rbr*{\xt{t+1}-\xt{t}}}dz\\
     & \leq -\frac{1}{\alpha}\norm{\xt{t+1}-\xt{t}}^2+\frac{1}{n}\sum_{i=1}^n\int_{z=0}^1\norm{ \rb{\nabla^2f_i(\xt{t}+z\vv_i) - \nabla^2f_i(\xt{t})}\uu_i}\norm{\xt{t+1}-\xt{t}}dz\\
     & \leq -\frac{1}{\alpha}\norm{\xt{t+1}-\xt{t}}^2+\frac{1}{n}\sum_{i=1}^n\int_{z=0}^1 \rho z\norm{\vv_i}^2\norm{\xt{t+1}-\xt{t}}dz\\
     & = -\frac{1}{\alpha}\norm{\xt{t+1}-\xt{t}}^2+\rho \frac{\beta^2}{2}\rb{\sum_{i=1}^n \frac{1}{n}\norm{\nabla f_i(\xt{t})-\nabla f(\xt{t})}^2}\norm{\xt{t+1}-\xt{t}}\\
     & = -\frac{1}{\alpha}\norm{\xt{t+1}-\xt{t}}^2+\rho \beta^2r(\xt{t})\norm{\xt{t+1}-\xt{t}}.
\end{align*}
Where the last two inequalities follow from Cauchy-Schwartz and $\rho$-Lipschitzness of $\nabla^2f_i$ respectively.

We can further use Equation \eqref{eq:taylorerror} to lower bound $\norm{\xt{t+1}-\xt{t}}$ as follows:
\begin{align*}
    \norm{\xt{t+1}-\xt{t}} &= \norm{-\alpha\nabla \hat{f}(\xt{t}) - \alpha \frac{1}{n}\sum_{i=1}^n \int_{z=0}^1 \rb{\nabla^2f_i(\xt{t}+z\vv_i) - \nabla^2f_i(\xt{t})}\vv_idz}\\
    &\geq  \norm{\alpha\nabla \hat{f}(\xt{t})} - \frac{1}{n}\sum_{i=1}^n\norm{ \alpha  \int_{z=0}^1 \rb{\nabla^2f_i(\xt{t}+z\vv_i) - \nabla^2f_i(\xt{t})}\vv_idz}\\
    &\geq  \norm{\alpha\nabla \hat{f}(\xt{t})} - \frac{1}{n}\sum_{i=1}^n\alpha  \int_{z=0}^1 \norm{\rb{\nabla^2f_i(\xt{t}+z\vv_i) - \nabla^2f_i(\xt{t})}\vv_i}dz\\
    &\geq \norm{\alpha\nabla \hat{f}(\xt{t})} - \frac{1}{n}\sum_{i=1}^n\alpha  \int_{z=0}^1 \rho\norm{z\vv_i}\norm{\vv_i}dz\\
    &=\norm{\alpha\nabla \hat{f}(\xt{t})} - \alpha\frac{\rho}{2} \rb{\sum_{i=1}^n \frac{1}{n}\norm{\vv_i}^2}\\
    &= \norm{\alpha\nabla \hat{f}(\xt{t})} - \alpha\rho \frac{\beta^2}{2}\rb{\sum_{i=1}^n \frac{1}{n}\norm{\nabla f_i(\xt{t})-\nabla f(\xt{t})}^2}\\
    &=\norm{\alpha\nabla \hat{f}(\xt{t})} - \alpha\rho \beta^2r(\xt{t}).
\end{align*}

Substituting in \eqref{eq:smoothFedGA}, we obtain:
\begin{align*}
    \hat{f}(\xt{t+1}) - \hat{f}(\xt{t}) &\leq  -\frac{1}{\alpha}\norm{\xt{t+1}-\xt{t}}^2+\rho \beta^2r(\xt{t})\norm{\xt{t+1}-\xt{t}}\\
    &+ \frac{L_1}{2}\norm{\xt{t+1}-\xt{t}}^2 + \frac{\beta L_2}{2}\norm{\xt{t+1}-\xt{t}}^2.
\end{align*}

To ensure the negativity of the coefficient for $\norm{\xt{t+1}-\xt{t}}^2$, we choose $\beta < \frac{L_1}{L_2}$.
Then, for $\alpha \leq \frac{1}{2L_1}$, we have:
\begin{equation}\label{eq:modified_desc}
    \hat{f}(\xt{t+1}) - \hat{f}(\xt{t}) \leq  -L_1\norm{\xt{t+1}-\xt{t}}^2+\rho \beta^2r(\xt{t})\norm{\xt{t+1}-\xt{t}}
\end{equation}

Thus a sufficient condition for $\hat{f}(\xt{t+1}) - \hat{f}(\xt{t}) < 0$ is:
\begin{equation}\label{eq:desc_cond}
    -L_1\norm{\xt{t+1}-\xt{t}}^2+\rho \beta^2r(\xt{t})\norm{\xt{t+1}-\xt{t}} < 0,
\end{equation}
or equivalently, 
\begin{align*}
    \beta^2 < \frac{L_1}{\rho}\frac{\norm{\xt{t+1}-\xt{t}}}{r(\xt{t})}.
\end{align*}
We now consider the following cases:
\begin{enumerate}
    \item \textbf{$\norm{\nabla{f(\xt{t})}}>0$}: Since $\lim_{\beta \rightarrow 0}\frac{L_1}{\rho}\frac{\norm{\xt{t+1}-\xt{t}}}{r(\xt{t})} =\frac{L_1\alpha}{\rho}\frac{\norm{\nabla{f(\xt{t})}}}{ r(\xt{t})} > 0$, $\exists \beta'$ such that $-L_1\norm{\xt{t+1}-\xt{t}}^2+\rho \beta'^2r(\xt{t})\norm{\xt{t+1}-\xt{t}} < 0$.
    \item $\norm{\nabla{f(\xt{t})}} = 0$ and $\norm{\nabla{r(\xt{t})}} > 0$. Then
    \begin{align*}
        \frac{L_1}{\rho}\frac{\norm{\xt{t+1}-\xt{t}}}{r(\xt{t})} &= \frac{L_1}{\rho}\frac{\norm{-\alpha\beta\nabla r(\xt{t}) - \alpha \frac{1}{n}\sum_{i=1}^n \int_{z=0}^1 \rb{\nabla^2f_i(\xt{t}+z\vv_i) - \nabla^2f_i(\xt{t})}\vv_idz}}{r(\xt{t})}\\
        &\geq \frac{L_1}{\rho}\frac{\norm{\alpha\beta\nabla r(\xt{t})} -\norm{\alpha \frac{1}{n}\sum_{i=1}^n \int_{z=0}^1 \rb{\nabla^2f_i(\xt{t}+z\vv_i) - \nabla^2f_i(\xt{t})}\vv_idz}}{r(\xt{t})}\\
        &\geq \frac{\alpha L_1}{\rho}\rbr*{\frac{\beta\norm{\nabla r(\xt{t})} -\rho\beta^2r(\xt{t})}{r(\xt{t})}}\\
    \end{align*}
    Thus it is sufficient to use $\beta'$ satisfying:
    \begin{align*}
        \beta'^2 \leq  \frac{\alpha L_1}{\rho}\rbr*{\frac{\beta'\norm{\nabla r(\xt{t})} -\rho\beta'^2r(\xt{t})}{r(\xt{t})}},
    \end{align*}
    or equivalently,
    \begin{align*}
        \beta' \leq \frac{\alpha L_1}{\rho\rbr{1+\alpha L_1}}\frac{\norm{\nabla r(\xt{t})} }{r(\xt{t})}.
    \end{align*}
\end{enumerate}
Combining with the assumption, $\beta < \frac{L_1}{L_2}$, we observe that in both cases, to ensure $\hat{f}(\xt{t+1}) - \hat{f}(\xt{t}) < 0$, it is sufficient to use $\beta$ satisfying:
\begin{align*}
   \beta < \min\cbr[bigg]{\beta',\frac{L_1}{L_2}}\\
\end{align*}
\end{proof}

\subsubsection{Implications of Theorem 3 for Convergence}

Since the aim of our proposed regularization is to encourage convergence to optima having better generalization properties, its utility is primarily for non-convex objectives, with multiple optima and critical points. Thus we prove convergence to a critical point of the regularized objective for smooth non-convex objectives under the descent condition derived in Theorem 3. While our assumptions appear quite strong, our empirical results show that they hold in practice.
Let $\beta_t$ denote the upper bound in the above derivation corresponding to the timestep $t$. Suppose there exists a $\beta$ satisfying the following condition:
\begin{equation}
    \beta^2 \leq (1-\epsilon)\beta^2_t,
\end{equation}

for some constant $\epsilon > 0$ independent of t. Then Equation \ref{eq:modified_desc} implies that:

\begin{align*}
    \hat{f}(\xt{t+1}) - \hat{f}(\xt{t}) &\leq  -L_1\norm{\xt{t+1}-\xt{t}}^2+\rho \beta^2r(\xt{t})\norm{\xt{t+1}-\xt{t}}\\
    &\leq -L_1\norm{\xt{t+1}-\xt{t}}^2+\rho \beta_t^2r(\xt{t})\norm{\xt{t+1}-\xt{t}}-\epsilon\rho\beta^2_t r(\xt{t})\norm{\xt{t+1}-\xt{t}}\\
    &\leq -\rho\epsilon\beta^2_t r(\xt{t})\norm{\xt{t+1}-\xt{t}}\\
    &\leq -\epsilon L_1\norm{\xt{t+1}-\xt{t}}^2.
\end{align*}
Where the last inequality follows from Equation \ref{eq:desc_cond}.
Re-arranging, we have:
\begin{align*}
    \epsilon L_1\norm{\xt{t+1}-\xt{t}}^2 &\leq \hat{f}(\xt{t+1}) - \hat{f}(\xt{t})\\
\end{align*}
After telescoping and averaging, we obtain:
\begin{align*}
 \frac{1}{T}\sum_{i=1}^T  L_1\norm{\xt{t+1}-\xt{t}}^2 \leq \frac{1}{\epsilon}\frac{1}{L_1 T}\hat{f}(\xt{0})
\end{align*}
Thus the mean size of updates, i.e. $\norm{\xt{t+1}-\xt{t}}^2$ converges at a rate of $\mathcal{O}(\frac{1}{T})$, analogous to the convergence rate of the mean squared gradient norm for smooth non-convex objectives.
\subsection{Linear Scaling}\label{app:linear}
The linear scaling rule \citep{goyal2018accurate}, when applied to a given (multi)set of $K$ minibatches $A$, proposes scaling the step size by $K$, while taking a gradient step on the combined objective $f_A(\xx)=\frac{1}{K}\sum_{i=1}^K\nabla f_{a_i}(\xx)$. As explained by \citet{goyal2018accurate}, a single scaled gradient step approximates $K$ SGD steps on the sequence of minibatches, since $-K\alpha\nabla f_A(\xx) = -\sum_{i=1}^K\alpha\nabla f_{a_i}(\xx)$. Using Lemma \ref{lemma:displacement}, we observe that $-K\alpha\nabla f_A(\xx)$ only incorporates the first order terms in $-\sum_{i=1}^K\alpha\nabla f_{a_i}(\xx)$. To incorporate the second order terms within a single update using the scaled step-size $K\alpha$, we require utilizing the displacement for each minibatch $a_i$ equal to the expectation of the displacement prior to the gradient step on $a_i$ conditioned on the given (multi)set $A$. Using the symmetry w.r.t time reversal, the expected displacement, upto the first order terms in $\alpha$, $\Ea{\vv_{a_i}}$ can be expressed as follows:
\begin{align*}
    \Ea{\vv_{a_i}} = -\frac{\alpha}{2}\rb{\sum_{j \neq i, j=1}^K \nabla f_{a_j}(\xx)}+\cO(\alpha^2).
\end{align*}
Thus the single-step approximation of SGD, with a linearly scaled step size $K\alpha$ is given by:
\begin{align*}
    \xx  \gets \xx - \alpha \sum_{i=1}^K \nabla f_{a_i}(\xx-\frac{\alpha}{2}\rb{\sum_{j \neq i, j=1}^K \nabla f_{a_j}(\xx)}).
\end{align*}
However, a major drawback of the above approximation is that for large $K$, the increase in step size amplifies the errors in the Taylor's theorem-based approximation for each gradient step. Therefore, to accurately assess the validity and effectiveness of the Taylor's theorem-based implicit regularization, we design algorithms GradAlign and FedGA compatible with small step sizes and arbitrarily large batches.
\subsection{Main Assumptions}\label{app:assumptions}
For Theorems \ref{thm:SGD},\ref{thm:GA},\ref{thm:FedGA regularization},\ref{thm:Scaffold regularization}, and the starting parameters $\xx$ under consideration, we assume that within a neighbourhood of $\xx$, the following conditions are satisfied: differentiability of $f_i(\cdot) \ \forall i$, differentiability of $r(\cdot)$ and  $\rho$-Lipschitzness of $\nabla^2f_i$ for some $\rho > 0$. We use the big-O notation $p(\beta) = \cO(q(\beta))$ for a positive scalar $\beta$ to represent the boundedness of $p(\beta)$ by $q(\beta)$ as $\beta \rightarrow 0$ i.e  $p(\beta) = \cO(q(\beta))$ implies that $\exists \beta' > 0$ such that $\abs{p(\beta)} \leq C\abs{q(\beta)}$ for all $0 \leq \beta \leq \beta'$ for some positive constant $C$. 
\subsection{Main Proofs}\label{app:proofs}

\subsubsection{Lemma 1}

\begin{proof}
By applying the fundamental theorem of calculus to each component of $f_i$, we obtain:
\begin{align*}
     \nabla f_i (\xx+\vv_{\xx}) &=  \nabla f_i (\xx) + \nabla^2 f_i (\xx)\vv_{\xx} + \int_{z=0}^1 \rb{\nabla^2f_i(\xx+z\vv_i) - \nabla^2f_i(\xx)}\vv_i dz.
\end{align*}
We bound the norm of the error term as follows:
\begin{align*}
    \norm{\nabla f_i (\xx+\vv_{\xx}) - \rb{\nabla f_i (\xx) + \nabla^2 f_i (\xx)\vv_{\xx}}} &= \norm{\int_{z=0}^1 \rb{\nabla^2f_i(\xx+z\vv_i) - \nabla^2f_i(\xx)}\vv_i dz}\\
     &\leq \int_{z=0}^1 \norm{\rb{\nabla^2f_i(\xx+z\vv_i) - \nabla^2f_i(\xx)}\vv_i} dz\\
     & \leq \int_{z=0}^1 \norm{\rho\norm{z\vv_i}\vv_i} dz\\
     &= \frac{\rho}{2}\norm{\vv_i}^2.
\end{align*}
Where the last inequality follows from the $\rho$-Lipschitzness of $\nabla^2f_i$.
\end{proof}

\subsubsection{Theorem \ref{thm:SGD}: SGD over K Sequential Steps}
\begin{proof}
The distribution over the sequences of $K$ steps, conditioned on the (multi)set $A=\{a_i\}_{i=1}^K$ of the sampled minibatches can be described through the corresponding distribution over re-orderings of $\{a_i\}_{i=1}^K$. We denote a randomly sampled re-ordering of $A$ as $A' = \{a'_i\}_{i=1}^K$, and the corresponding reverse ordering by $A'_{-1}$. The symmetry w.r.t time-reversal implies that the probability distribution $P$ over $A'$ satisfies $P(A') = P(A'_{-1})$. For a sequence of SGD steps under a given ordering $A'$, we denote by $g_{A',i}(\xx)$, the $i_{th}$ gradient step corresponding to $A'$ and the starting parameters $\xx$ and the displacement from the starting point $\xx$ prior to the $i_{th}$ gradient step by $\vv_{A'}^{(i)}(\xx)$. Similarly, we denote the $i_{th}$ gradient step and the corresponding displacement for $K$ sequential gradient steps on the mean objective by $g_{GD}^{(i)}(\xx)$ and $\vv_{GD}^{(i)}(\xx)$.
Using Lemma \ref{lemma:displacement}, we have:
\begin{equation}\label{eq:g_SGD}
\begin{split}
    g_{A'}^{(i)}(\xx) &= -\alpha\nabla f_{a'_i} (\xx+\vv_{A'}^{(i)}(\xx)) =  -\alpha\rb{\nabla f_{a'_i} (\xx) +\nabla^2 f_{a'_i} (\xx)\vv_{A'}^{(i)}(\xx) + \cO(\norm{\vv_{A'}^{(i)}(\xx)}^2)}\\
    &=-\alpha\nabla f_{a'_i} (\xx) - \alpha\nabla^2 f_{a'_i} (\xx)\vv_{A'}^{(i)}(\xx) + \alpha\cO(\norm{\vv_{A'}^{(i)}(\xx)}^2).
\end{split}
\end{equation}
Where $\vv_{A'}^{(i)}(\xx) = \sum_{j=1}^{i-1} g_{A'}^{(j)}(\xx)$. For $i=2$, we obtain:
\begin{align*}
    g_{A'}^{(2)}(\xx) &= -\alpha\rb{\nabla f_{a'_2} (\xx)+\nabla^2 f_{a'_1} (\xx)\nabla f_{a'_1}(\xx)+ \cO(\norm{\alpha\nabla f_{a'_1}(\xx)}^2)}\\
    &= -\alpha\nabla f_{a'_2} (\xx)+ \alpha^2\nabla f_{a'_1} (\xx)\nabla f_{a'_1}(\xx)+\cO(\alpha^3).
\end{align*}
By applying Equation \eqref{eq:g_SGD} inductively for $i=3,\dots,K$, we obtain:
\begin{equation}\label{eq:SGD_v}
\begin{split}
   \vv_{A'}^{(i)}(\xx) &= \sum_{j=1}^{i-1} g_{A'}^{(j)}(\xx)\\
    &= \sum_{j=1}^{i-1}-\alpha\nabla f_{a'_j}(\xx)-\alpha\nabla^2 f_{a'_j} (\xx)\vv_{A'}^{(j)}(\xx)+\cO(\alpha^3)\\
    &= \sum_{j=1}^{i-1}-\alpha\nabla f_{a'_j}(\xx)-\alpha\nabla^2 f_{a'_j} (\xx)\rb{-\alpha\rb{\sum_{l=1}^{j-1} g_{A'}^{(l)}(\xx)}}+\cO(\alpha^3)\\
    &= \sum_{j=1}^{i-1}-\alpha\nabla f_{a'_j}(\xx)+\cO(\alpha^2),\\
\end{split}
\end{equation}
and as a result:\begin{equation}\label{eq:SGD_g}
\begin{split}
    g_{A'}^{(i)}(\xx) &= -\alpha\nabla f_{a'_i}(\xx)-\alpha\nabla^2 f_{a'_i} (\xx)\rb{\vv_{A'}^{(i)}(\xx)}+\cO(\alpha^3)\\
    &= -\alpha\nabla f_{a'_i} (\xx)+ \alpha\nabla^2 f_{a'_1} (\xx)\rb{\sum_{j=1}^{i-1}\nabla f_{a'_j}(\xx)}+\cO(\alpha^3).
\end{split}
\end{equation}

Similarly, for gradient descent on the mean objective, we have:
\begin{align*}
    \vv_{GD}^{(i)}(\xx)&= \rb{\sum_{j=1}^{i-1} g_{GD}^{(j)}(\xx)}\\
    &= \sum_{j=1}^{i-1}-\alpha\nabla f_{A}(\xx)-\alpha\nabla^2 f_{A} (\xx)\vv_{A}^{(j)}(\xx)+\cO(\alpha^3)\\
    &=  \sum_{j=1}^{i-1}-\alpha\nabla f_{A}(\xx)-\alpha\nabla^2 f_{A} (\xx)\rb{-\alpha\rb{\sum_{l=1}^{j-1} g_{GD}^{(l)}(\xx)}}+\cO(\alpha^3)\\
    &= \sum_{j=1}^{i-1}-\alpha\nabla f_{A}(\xx)+\cO(\alpha^2).
\end{align*}
Therefore, the $i_{th}$ gradient step for gradient descent on the mean objective is given by:
\begin{align*}
    g_{GD}^{(i)}(\xx) &= -\alpha\nabla f_{A} (\xx)-\alpha\nabla^2 f_{A} (\xx)\vv_{GD}^{(i)}(\xx)+\cO(\alpha^3)\\
    &= -\alpha\nabla f_{A} (\xx)+ \alpha^2\nabla f_{A} (\xx)\rb{\sum_{j=1}^{i-1}\nabla f_{A}(\xx)}+\cO(\alpha^3)\\
    &= -\alpha\nabla f_{A} (\xx)+ \alpha^2\nabla f_{A} (\xx)\rb{(i-1)\nabla f_{A}(\xx)}+\cO(\alpha^3).
\end{align*}
The expected difference between the parameters reached after $K$ steps of SGD using the corresponding mini-batches in $A$ and $K$  steps of GD on the mean objective $f_A(\xx) = \frac{1}{K}\sum_{i=1}^K f_{a_i}(\xx)$ with initial parameters $\xx$ is then given by:

\begin{align*}
 &\Eb{A'}{\sum_{i=1}^{K} (g_{A'}^{(i)}(\xx) - g_{GD}^{(i)}(\xx))}\\
    =& \Eb{A'}{-\alpha\nabla f_{a'_i} (\xx)+ \alpha\nabla^2 f_{a'_1} (\xx)\rb{\sum_{j=1}^{i-1}\nabla f_{a'_j}(\xx)}+\cO(\alpha^3)}\\ 
    +& \Eb{A'}{\sum_{i=1}^K\alpha\nabla f_{A} (\xx)+ \alpha^2\nabla f_{A} (\xx)\rb{(i-1)\nabla f_{A}(\xx)}+\cO(\alpha^3)}\\
    =& \sum_{A' \in S_K} P(A')  \left(\alpha^2(\sum_{i=1}^{K}\sum_{j=1}^{i-1} \nabla^2 f_{a'_i}(\xx) \nabla_{a'_j} f(\xx)) - \alpha\frac{K(K-1)}{2}\nabla^2 f(\xx)_A \nabla f_{A}(\xx)  +\cO(\alpha^3) \right)\\
    =& \frac{1}{2}\left(\sum_{A \in [m]^K} P(A')  \left(\alpha^2(\sum_{i=1}^{K}\sum_{j=1}^{i-1} \nabla^2 f_{a'_i}(\xx) \nabla_{a'_j} f(\xx)) - \alpha\frac{K(K-1)}{2}\nabla^2 f(\xx)_A \nabla f_{A}(\xx) +\cO(\alpha^3) \right) \right.\\
    &+  \left. \sum_{A \in [m]^K} P(A'_{-1})  \left(\alpha^2(\sum_{i=1}^{K}\sum_{j=1}^{i-1} \nabla^2_{a'_{K+1-i}} f(\xx) \nabla_{a_{K+1-j}} f(\xx)) - \alpha\frac{K(K-1)}{2}\nabla^2 f(\xx)_A \nabla f_{A}(\xx)  +\cO(\alpha^3) \right)\right)\\
    =& \sum_{A' \in S_K} P(A') \left(\alpha^2(\sum_{i=1}^{K}\sum_{j=1}^{K} \nabla^2 f_{a'_i}(\xx) \nabla_{a'_j} f(\xx)) - \frac{\alpha^2}{2}(\sum_{i=1}^{K}\sum_{j=1}^{K} \nabla^2 f_{a'_i}(\xx) \nabla f_{a'_j}(\xx)) +  \alpha^2\frac{K}{2}\nabla^2 f_A(\xx) \nabla f_{A}(\xx) +\cO(\alpha^3) \right).
\end{align*}
Now, since each $A'$ corresponds to a re-ordering of the given (multi)set $A$, the above expression simplifies to:
\begin{align*}
 &\Eb{A'}{\sum_{i=1}^{K} (g_{A'}^{(i)}(\xx) - g_{GD}^{(i)}(\xx))}\\
    =& \alpha^2\frac{K^2}{2}\nabla^2 f_{A}(\xx) \nabla f_{A}(\xx) - \frac{\alpha^2}{2}\sum_{i=1}^{K} \nabla^2 f_{a_i}(\xx) \nabla f_{a_i}(\xx) - \alpha^2\frac{K^2}{2}\nabla^2 f_{A}(\xx) \nabla f_{A}(\xx) +  \alpha^2\frac{K}{2}\nabla^2 f_{A}(\xx) \nabla f_{A}(\xx) + \cO(\alpha^3)\\
    =& -\Eb{A'}{\frac{\alpha^2}{2} ( \sum_{i=1}^{K} (\nabla^2 f_{a_i}(\xx) \nabla f_{a_i}(\xx) - \nabla^2 f_{a_i}(\xx) \nabla f_{A}(\xx) - \nabla^2 f_{A}(\xx) \nabla f_{a_i}(\xx) + \nabla^2 f_{A}(\xx) \nabla f(\xx)))} + \cO(\alpha^3)\\
    =& -\frac{\alpha^2}{4} (\sum_{i=1}^{K}(\nabla^2 f_{a_i}(\xx)-\nabla^2 f_{A}(\xx))(\nabla f_{a_i}(\xx)-\nabla f_{A}(\xx))) +\cO(\alpha^3)\\
    =& -\frac{\alpha^2}{4}(\sum_{i=1}^{K}(\nabla^2 f_{a_i}(\xx)-\nabla^2 f_{A}(\xx))(\nabla f_{a_i}(\xx)-\nabla f_{A}(\xx)))+\cO(\alpha^3)\\
    =& -\frac{\alpha^2}{4}\nabla_{\xx}\Big(\sum_{i=1}^{K} \|\nabla f_{a_i}(\xx)-\nabla f_{A}(\xx)\|^2\Big) = -\frac{K\alpha^2}{2}{\nabla r_A(\xx)}+\cO(\alpha^3).
\end{align*}
\end{proof}

\subsubsection{Approximating K SGD steps with K GD steps on the Regularized Objective}

In this section, we prove that for a given sequence $A$ of $K$ minibatches, the expected difference between $K$ updates using SGD and gradient descent on the mean objective (Equation \eqref{eq:regularization}) can be approximated through gradient descent on the regularized mean objective $\hat{f}_A(\xx) =f_A(\xx)+\frac{\alpha}{2}r_A(\xx)$
Similar to the proof for Theorem \ref{thm:SGD}, for a given sequence $A$, we denote the $i_{th}$ gradient step and the displacement from $\xx$ prior to it under the mean objective $f_A(\xx)$ and the regularized mean objective $\hat{f}_A(\xx)$ by $g_{GD}^{(i)}(\xx),\vv_{GD}^{(i)}(\xx)$ and  $\hat{g}_{GD}^{(i)}(\xx),\hat{\vv}_{GD}^{(i)}(\xx)$ respectively.

We have:
\begin{equation}\label{eq:g_Ghat}
\begin{split}
    \hat{g}_{GD}^{(i)} &= -\alpha\nabla \hat{f}_{A} (\xx+\hat{\vv}_{GD}^{(i)}(\xx)) = -\alpha\rb{\nabla f_A(\xx+\hat{\vv}_{GD}^{(i)}(\xx))+\frac{\alpha}{2}\nabla r(\xx+\hat{\vv}_{GD}^{(i)}(\xx))}\\
    &=  -\alpha\rb{\nabla f_{A} (\xx) +\nabla^2 f_{A} (\xx)\hat{\vv}_{GD}^{(i)}(\xx) + \cO(\norm{\hat{\vv_{GD}^{(i)}}}^2) +\frac{\alpha}{2} \nabla r(\xx) -\frac{\alpha}{2}\nabla^2 r(\xx)\hat{\vv}_{GD}^{(i)}(\xx) + \cO(\norm{\hat{\vv}_{GD}^{(i)}}^2)} \\
\end{split}
\end{equation}
For $i=2$, we get:
\begin{align*}
    \hat{g}_{GD}^{(2)}
    &=  -\alpha\rb{\nabla f_{A} (\xx) -\alpha\nabla^2 f_{A} (\xx)\rb{\nabla f_A(\xx)+\frac{\alpha}{2}\nabla r_A(\xx)} + \cO(\alpha^2)}\\
    &-\alpha\rb{\frac{\alpha}{2} \nabla r(\xx) -\frac{\alpha^2}{2}\nabla^2 r(\xx)\rb{\nabla f_A(\xx)+\frac{\alpha}{2}\nabla r_A(\xx)} + \cO(\alpha^2)}\\
    &-\alpha\nabla f_{A} (\xx) +\alpha^2\nabla^2 f_{A} (\xx)\nabla f_A(\xx))-\frac{\alpha^2}{2} \nabla r(\xx) + \cO(\alpha^3).
\end{align*}
By inductively applying Equation \eqref{eq:g_Ghat} for $i=3,\cdots,K$, we obtain:
\begin{align*}
    \hat{\vv}_{GD}^{(i)}(\xx)&= \rb{\sum_{j=1}^{i-1} \hat{g}_{GD}^{(j)}(\xx)}\\
    &= \sum_{j=1}^{i-1}-\alpha\nabla f_{A}(\xx)-\alpha\nabla^2 f_{A} (\xx)\vv_{A}^{(j)}(\xx)+\cO(\alpha^3)\\
    &=  \sum_{j=1}^{i-1}-\alpha\nabla f_{A}(\xx)-\alpha\nabla^2 f_{A} (\xx)\rb{-\alpha\rb{\sum_{l=1}^{j-1} g_{GD}^{(l)}(\xx)}}+\cO(\alpha^3)\\
    &= \sum_{j=1}^{i-1}-\alpha\nabla f_{A}(\xx)+\cO(\alpha^2),
\end{align*}
and therefore,
\begin{align*}
    \hat{g}_{GD}^{(i)}(\xx) &= -\alpha\nabla \hat{f}_{A} (\xx+\hat{\vv}_{GD}^{(i)}(\xx))\\
    &= -\alpha\nabla f_{A} (\xx) -\alpha\nabla^2 f_{A} (\xx)\hat{\vv}_{GD}^{(i)}(\xx)-\frac{\alpha^2}{2} \nabla r(\xx)-\frac{\alpha^2}{2} \nabla^2 r(\xx)\hat{\vv}_{GD}^{(i)}(\xx)+\cO(\alpha^3)\\
    &=-\alpha\nabla f_{A} (\xx) +\alpha^2\nabla^2 f_{A} (\xx)\rb{\sum_{j=1}^{i-1}\nabla f_{A}(\xx)} -\frac{\alpha^2}{2} \nabla r(\xx)+\cO(\alpha^3).
\end{align*}
Thus the difference between the parameters reached by $K$ gradient descent steps on the regularized mean objective and the mean objective, denoted by $\hat{\xx}_{GD,A}$ and $\xx_{GD,A}$ respectively is given by: 
\begin{align*}
&\hat{\xx}_{GD,A}-\xx_{GD,A}=\sum_{i=1}^{K} \rb{\hat{g}_{GD}^{(i)}-g_{GD}^{(i)}}\\
&= \sum_{i=1}^{K} \rb{-\alpha\nabla f_{A} (\xx) +\alpha^2\nabla^2 f_{A} (\xx)\rb{\sum_{j=1}^{i-1}\nabla f_{A}(\xx)} -\frac{\alpha^2}{2} \nabla r(\xx)+\cO(\alpha^3)} \\
&-\rb{-\alpha\nabla f_{A} (\xx)+ \alpha^2\nabla f_{A} (\xx)\rb{(i-1)\nabla f_{A}(\xx)}+\cO(\alpha^3)}\\
&= \sum_{i=1}^{K}-\frac{\alpha^2}{2} \nabla r(\xx)+\cO(\alpha^3)\\
&=-\frac{\alpha^2K}{2} \nabla r(\xx)+\cO(\alpha^3).
\end{align*}

\subsubsection{Theorem \ref{thm:GA}: GradAlign}
\begin{proof}
Using Lemma \ref{lemma:displacement}, the gradient step $g_i(\xx)$ for the $i_{th}$ mini-batch obtained after displacement through  $\vv_i(\xx) = -\beta\rb{\nabla f(\xx)- \nabla f_i(\xx)}$ with starting parameters $\xx$, can be expressed as:

\begin{equation}\label{eq:g_GA}
\begin{split}
    g_{i} &= -\alpha\nabla f_i (\xx+\vv_i(\xx)) =  -\alpha\rb{\nabla f_i (\xx) +\nabla^2 f_i (\xx)\vv_i(\xx) + \cO(\norm{\vv_i}^2)}\\
    &= -\alpha\rb{\nabla f_i (\xx) -\beta\nabla^2 f_i (\xx)\rb{\nabla f(\xx)- \nabla f_i(\xx)} + \cO(\norm{\beta\rb{\nabla f(\xx)- \nabla f_i(\xx)}}^2)}\\
    &=-\alpha\nabla f_i (\xx) +\alpha\beta\nabla^2 f_i (\xx)\rb{\nabla f(\xx)- \nabla f_i(\xx)} + \cO(\alpha\beta^2).
\end{split}
\end{equation}
Therefore, we obtain:
 \begin{align*}
    &\xx_{GA} - \xx_{GD} \\
    &= -\frac{\alpha}{n} \sum_{i=1}^n \nabla f_i(\xx) +\frac{\alpha}{n}\sum_{i=1}^n \beta\nabla^2 f_i(\xx)\rb{\nabla f(\xx) -\nabla f_i(\xx)}  +\cO(\alpha\beta^2) +\frac{\alpha}{n} \sum_{i=1}^n \nabla f_i(\xx)\\
     & -\frac{\alpha\beta}{n} ( \sum_{i=1}^{n} (\nabla^2 f_i(\xx) \nabla f_i(\xx) - \nabla^2 f(\xx) \nabla f(\xx))) + \cO(\alpha\beta^2)\\
    =& -\frac{\alpha\beta}{n} ( \sum_{i=1}^{n} (\nabla^2 f_i(\xx) \nabla f_i(\xx) - \nabla^2 f_i(\xx) \nabla f(\xx) - \nabla^2 f(\xx) \nabla f_i(\xx) + \nabla^2 f(\xx) \nabla f(\xx))) + \cO(\alpha\beta^2)\\
    =& -\frac{\alpha\beta}{n} (\sum_{i=1}^{n}(\nabla^2 f_i(\xx)-\nabla^2 f(\xx))(\nabla f_i(\xx)-\nabla f(\xx)))) +\cO(\alpha\beta^2)\\
    =& -\frac{\alpha\beta}{2n}\nabla_{\xx}((\sum_{i=1}^{n} \|\nabla f_i(\xx)-\nabla f(\xx)\|^2) +\cO(\alpha\beta^2).
 \end{align*}
\end{proof}

\subsubsection{Theorem \ref{thm:FedGA regularization}: FedGA}

\begin{proof}
 Analogous to the proof for \ref{thm:SGD}, we denote the local displacement for client $i$ from the starting point $\xx$ prior to the $k_{th}$ step for FedAvg and FedGA by $\vv_{i,FedAvg}^{(k)},\vv_{i,FedGA}^{(k)}$ respectively and the corresponding $k_{th}$ gradient step by $g_{i,FedAvg}^{(k)}(\xx),g_{i,FedGA}^{(k)}(\xx)$ respectively.
 For a given client $i$, the $K$ local updates in FedAvg are equivalent to a sequnce of SGD updates on the sampled $K$ minibatches. Thus, using Equations \eqref{eq:SGD_v},\eqref{eq:SGD_g}, we have:
\begin{align*}
   \vv_{i,FedAvg}^{(k)}(\xx) &= \sum_{j=1}^{k-1} g_{i,FedAvg}^{(j)}(\xx)\\
    &= \sum_{j=1}^{k-1}-\alpha\nabla f_{i}(\xx;\zeta_{i,j})-\alpha\nabla^2 f_{i} (\xx;\zeta_{i,j})\vv_{i}^{(j)}(\xx)+\cO(\alpha^3)\\
    &= \sum_{j=1}^{k-1}-\alpha\nabla f_{i}(\xx;\zeta_{i,j})-\alpha\nabla^2 f_{i} (\xx;\zeta_{i,j})\rb{\sum_{l=1}^{j-1} g_{i,FedAvg}^{(l)}(\xx)}+\cO(\alpha^3)\\
    &= \sum_{j=1}^{i-1}-\alpha\nabla f_{i}(\xx;\zeta_{i,j})+\cO(\alpha^2),
\end{align*}
and
\begin{align*}
    g_{i,FedAvg}^{(k)}(\xx) &= -\alpha\nabla f_{i}(\xx;\zeta_{i,k})-\alpha\nabla^2 f_{i} (\xx)\rb{\vv_{i,FedAvg}^{(k)}(\xx)}+\cO(\alpha^3)\\
    &= -\alpha\nabla f_{i} (\xx;\zeta_{i,k})+ \alpha\nabla^2 f_{i} (\xx;\zeta_{i,k})\rb{\sum_{j=1}^{k-1}\nabla f_{i}(\xx;\zeta_{i,j})}+\cO(\alpha^3).
\end{align*}
Whereas for FedGA, we include an additional gradient alignment displacement $-\beta\rb{\nabla f(\xx)- \nabla f_i(\xx)}$ for each local update to obtain: 
\begin{align*}
   \vv_{i,FedGA}^{(k)}(\xx) &= -\beta\rb{\nabla f(\xx)- \nabla f_i(\xx)}+\sum_{j=1}^{k-1} g_{i,FedGA}^{(j)}(\xx)\\
   &= -\beta\rb{\nabla f(\xx)- \nabla f_i(\xx)}-\sum_{j=1}^{k-1}\alpha\nabla f_{i}(\xx;\zeta_{i,j})  +\cO(\alpha^2)+\cO(\alpha\beta)
\end{align*}
and
\begin{align*}
    &g_{i,FedGA}^{(k)}(\xx) = -\alpha\nabla f_{i}(\xx;\zeta_{i,k})-\alpha\nabla^2 f_{i} (\xx)\rb{\vv_{i,FedGA}^{(k)}(\xx)}+\cO(\alpha^3)+\cO(\alpha\beta^2)\\
    &= -\alpha\nabla f_{i} (\xx;\zeta_{i,k})- \alpha\nabla^2 f_{i} (\xx;\zeta_{i,k})\rb{-\beta\rb{\nabla f(\xx)- \nabla f_i(\xx)}-\sum_{j=1}^{k-1}\alpha\nabla f_{i}(\xx;\zeta_{i,j})}+\cO(\alpha^3) +\cO(\alpha\beta^2).
\end{align*}
The expected difference between the parameters obtained after one round of FedGA and FedAvg is then given by:
\begin{align*}
     \Ea{\xx_{FedGA} - \xx_{FedAVG}} &= \Ea{\frac{1}{n}\sum_{i=1}^n\sum_{k=1}^K g_{i,FedGA}^{(k)}(\xx) -\frac{1}{n}\sum_{i=1}^n\sum_{k=1}^Kg_{i,FedAvg}^{(k)}(\xx)}.
\end{align*}
Where the expectation is over random variables $\{\zeta_{i,k}\}_{k=1}^K$ controlling the stochasticity of the local updates for each client $i$.
Linearity of expectation allows us to couple the local updates for FedGA and FedAvg by using the same $\zeta_{i,k}$ for both the algorithms for each client $i$ and update $k$. We obtain:
\begin{align*}
    &\Ea{\xx_{FedGA} - \xx_{FedAVG}}
    = -\Ea{\frac{\alpha}{n} \sum_{i=1}^n \sum_{k=1}^K \nabla f_i(\xx;\zeta_{i,l})}\\ &+\Ea{\frac{\alpha}{n}\sum_{i=1}^n \sum_{k=1}^K \nabla^2 f_i(\xx;\zeta_{i,k})\rb{\beta \rb{\nabla f(\xx)-\nabla f_i(\xx)} + \alpha \sum_{l=1}^{k-1}\nabla f_i(\xx;\zeta_{i,l})}} +\cO(\alpha\beta^2) \\
    &-\Ea{(-\frac{\alpha}{n} \sum_{i=1}^n \sum_{k=1}^K\nabla f_i(\xx;\zeta_{i,k})  + \frac{\alpha}{n}\sum_{i=1}^n \sum_{k=1}^K \nabla^2 f_i(\xx;\zeta_{i,k})(\alpha \sum_{l=1}^{k-1}\nabla f_i(\xx;\zeta_{i,l}))}+\cO(\alpha\beta^2)\\
     &= Ea{\frac{\alpha\beta}{n} \sum_{i=1}^{n} \rb{\sum_{k=1}^K \nabla^2  f_i(\xx;\zeta_{i,k}) \rb{\nabla f(\xx) - \nabla f_i(\xx)})}} + \cO(\alpha\beta^2)\\
    =& -\frac{\alpha\beta K}{n} ( \sum_{i=1}^{n} (\nabla^2 f_i(\xx) \nabla f_i(\xx) - \nabla^2 f_i(\xx) \nabla f(\xx) - \nabla^2 f(\xx) \nabla f_i(\xx) + \nabla^2 f(\xx) \nabla f(\xx))) + \cO(\alpha\beta^2)\\
    =& -\frac{\alpha\beta K}{n} (\sum_{i=1}^{n}(\nabla^2 f_i(\xx)-\nabla^2 f(\xx))(\nabla f_i(\xx)-\nabla f(\xx)))) +\cO(\alpha\beta^2)\\
    =& -\frac{\alpha\beta K}{2n}\nabla_{\xx}\left(\sum_{i=1}^{n} \norm{\nabla f_i(\xx)-\nabla f(\xx)}^2\right) +\cO(\alpha\beta^2)\\
\end{align*}
\end{proof}
\subsection{Implicit cancellation in FedGA}\label{app:FedGA}
In this section, we describe the equivalence between using the displacement $-\beta\rb{\nabla f(\xx)- \nabla f_i(\xx)}$ only once at the beginning of each round for each client $i$ in FedGA, and using the same displacement, but on each of the $K$ local updates. The former version of the algorithm is described in Algorithm \ref{alg:FedGA} while the latter is described below in Algorithm~\ref{alg:FedGA2}.

\begin{algorithm}
  \caption{Federated  Gradient Alignment (FedGA)}%
  \label{alg:FedGA2}
  \begin{algorithmic}[1]{
      \State learning rate $\alpha$
      \State Initial model parameters :$\xx$
      \State Mean of initial gradients for clients in $[n]$: $\og= \frac{1}{n}\sum_{i=1}^n\nabla f_i(\xx)$
      \While{not done}
        \State{$\og\gets \frac{1}{n}\sum_{i=1}^{n}\nabla f_i(\xx)$} 
        \Comment{Update the mean gradient}
        \For{Client $i$ in $[1,\cdots,n]$}
          \State{Obtain the displacement of the mean gradient as $\vv_i \gets \rb{\og- \nabla f_i(\xx)}$}
          \State{$\xx^{(0)}_i \gets \xx$}
          \For{$k$ in $[1,\cdots,K]$}
          \State{$\xx^{(k)}_i \gets \xx^{(k-1)}_i -  \alpha\nabla{f_i}(\xx^{(k-1)}_i-\beta \vv_i;\zeta_{i,k})$}\label{step:update}
          \Comment{Obtain gradient after displacement}
          \EndFor
        \EndFor
      \State{$\xx \gets \frac{1}{n}\sum_{i=1}^{n}\xx^{(K)}_i$}
      \EndWhile
      }
    \end{algorithmic}
\end{algorithm}
 
Notice that to compute $\xx^{(k)}_i$ in line~\ref{step:update} of Algorithm~\ref{alg:FedGA2} we could instead follow these 3 steps: (1) $\xx^{(k)}_i \gets \xx^{(k)}_i + \vv_i$, then (2) $\xx^{(k)}_i \gets \xx^{(k)}_i - \alpha\nabla{f_i}(\xx^{(k)}_i)$, and finally $\xx^{(k)}_i \gets \xx^{(k)}_i -\vv_i$ to arrive at the same point obtained in line~\ref{step:update}. 
Since $\vv_i$ remains constant throughout the $K$ steps in one round, the displacement in step (1) and step (3) cancel between consecutive local updates. 
Thus, we are left with the first and last displacement only.
Furthermore, since the displacements average to $0$ i.e $\sum_{i=1}^n \vv_i = \sum_{i=1}^n  -\beta\rb{\nabla f(\xx)-\nabla f_i(\xx)} = 0$, we do not need to take the final step either, and hence we are left with the formulation of Algorithm~\ref{alg:FedGA}.

\subsection{SCAFFOLD}\label{app:scaffold}
The full SCAFFOLD algorithm \citep{pmlr-v119-karimireddy20a} is described in Algorithm \ref{alg:scaffold}. For simplicity, we assume that the displacement we use is computed only among the sampled clients.
\begin{algorithm}[h]
  \caption{Scaffold}%
  \label{alg:scaffold}
  \begin{algorithmic}[1]{
      \State learning rate $\alpha$
      \State Initial model parameters :$\xx$
      \While{not done}
        \State{$\og= \frac{1}{n}\sum_{i=1}^n\nabla f_i(\xx)$}
        \Comment{Compute each $\nabla f_i(\xx;)$ in parallel}
        \For{Client $i$ in $[1,\cdots,n]$}
          \State{$\xx^{(0)}_i \gets \xx$}
          \For{$k$ in $[1,\cdots,K]$}
          \State $\xx^{(k)}_i \gets \xx^{(k-1)}_i -  \alpha\left(\nabla{f_i}(\xx^{(k-1)}_i;\zeta_{i,k}) + \nabla f(\xx) - \nabla f_i(\xx) \right )$
          \EndFor
      \EndFor
      \State{$\xx \gets \frac{1}{n}\sum_{i=1}^{n}\xx^{(K)}_i$}
        \EndWhile
      }
    \end{algorithmic}
\end{algorithm}

We observe that unlike $FedGA$, the displacement $\vv_{i,SCAFFOLD}^{(k)}$ from the starting parameters $\xx$ prior to the $k_{th}$ gradient step for SCAFFOLD, involves $k-1$ drift correction terms $-\alpha \rb{\nabla f(\xx) - \nabla f_i(\xx)}$ in addition to the $k-1$ local gradient steps. Thus we have:
\begin{align*}
   \vv_{i,SCAFFOLD}^{(k)}(\xx) &= -(k-1)\alpha\rb{\nabla f(\xx)- \nabla f_i(\xx)}+\sum_{j=1}^{k-1} g_{i,SCAFFOLD}^{(j)}(\xx),
\end{align*}
where $g_{i,SCAFFOLD}^{(j)}(\xx)$ denotes the $j_{th}$ gradient step for client $i$. Smilar to FedGA and SGD, $g_{i,SCAFFOLD}^{(j)}(\xx)$ can be evaluated by inductively computing the local displacements and gradient steps to obtain:
\begin{align*}
    &g_{i,SCAFFOLD}^{(k)}(\xx) = -\alpha\nabla f_{i}(\xx;\zeta_{i,k})\\
    &-\alpha\nabla^2 f_{i} (\xx)\rb{\vv_{i,SCAFFOLD}^{(k)}(\xx)}+\cO(\alpha^3)\\
    &= -\alpha\nabla f_{i} (\xx;\zeta_{i,k})-\alpha\nabla^2 f_{i} (\xx;\zeta_{i,k})\rb{-(k-1)\alpha\rb{\nabla f(\xx)- \nabla f_i(\xx)}-\sum_{j=1}^{k-1}\alpha\nabla f_{i}(\xx;\zeta_{i,j})} +\cO(\alpha^3).
\end{align*}
The expected difference between the parameters obtained after one round of SCAFFOLD and FedAvg is then given by:
\begin{align*}
     &\Ea{\xx_{SCAFFOLD} - \xx_{FedAVG}}\\ 
     &= \Ea{\frac{1}{n}\sum_{i=1}^n\rb{\sum_{k=1}^K g_{i,SCAFFOLD}^{(k)}(\xx) -\alpha\rb{\nabla f(\xx)- \nabla f_i(\xx)}} -\frac{1}{n}\sum_{i=1}^n\sum_{k=1}^K g_{i,FedAvg}^{(k)}(\xx)} \\
     = &-\Ea{\frac{\alpha}{n} \sum_{i=1}^n \sum_{k=1}^K \rb{\nabla f_i(\xx;\zeta_{i,k})+\rb{\nabla f(\xx)- \nabla f_i(\xx)}}}\\ &+\Ea{\frac{\alpha}{n}\sum_{i=1}^n \sum_{k=1}^K \nabla^2 f_i(\xx;\zeta_{i,k})\rb{\alpha(k-1) \rb{  \nabla f(\xx)-\nabla f_i(\xx)} 
     + \alpha\sum_{l=1}^{k-1}\nabla f_i(\xx;\zeta_{i,l})}} +\cO(\alpha^3) \\
    &+\Ea{(\frac{\alpha}{n} \sum_{i=1}^n \sum_{k=1}^K\nabla f_i(\xx;\zeta_{i,k})  - \frac{\alpha}{n}\sum_{i=1}^n \sum_{k=1}^K \nabla^2 f_i(\xx;\zeta_{i,k})(\alpha \sum_{l=1}^{k-1}\nabla f_i(\xx;\zeta_{i,l})) +\cO(\alpha^3))} \\
     &= -\Ea{\frac{\alpha^2 K(K-1)}{2n} ( \sum_{i=1}^{n} (\nabla^2 f_i(\xx;\zeta_{i,l}) \rb{\nabla f_i(\xx) - \nabla f(\xx)}))} + \cO(\alpha^3)\\
    =& -\frac{\alpha^2 K(K-1)}{2n} ( \sum_{i=1}^{n} (\nabla^2 f_i(\xx) \nabla f_i(\xx) - \nabla^2 f_i(\xx) \nabla f(\xx) - \nabla^2 f(\xx) \nabla f_i(\xx) + \nabla^2 f(\xx) \nabla f(\xx))) + \cO(\alpha^3)\\
    =& -\frac{\alpha^2 K(K-1)}{2n} (\sum_{i=1}^{n}(\nabla^2 f_i(\xx)-\nabla^2 f(\xx))(\nabla f_i(\xx)-\nabla f(\xx)))) +\cO(\alpha^3)\\
    =& -\frac{\alpha^2 K(K-1)}{4n}\nabla_{\xx}((\sum_{i=1}^{n} \norm{\nabla f_i(\xx)-\nabla f(\xx)}^2) +\cO(\alpha^3).
\end{align*}

\paragraph{``Consistency'' and ``Efficiency''}: As explained in the previous section, FedGA only requires adding the displacement once for each worker, before taking the sequential steps. Following the application of Taylor's theorem using Lemma 1, we see that due to the application of a new step of displacement before each local step in SCAFFOLD, the second-order term responsible for gradient alignment across clients grows in magnitude as the number of local steps increases. This results in a larger difference between successive local steps and causes the inter-client alignment terms to be larger for the latter local steps than the initial ones. Whereas for FedGA, since a single displacement is applied initially, all local steps receive an inter-client alignment displacement of the same magnitude. We believe this is especially important in the presence of minibatch sampling within each client, since it leads to the same alignment effect for all local mini-batches. We refer to the above distinction between SCAFFOLD and FedGA as improvement in the ``consistency'' while not requiring the addition of the displacement at each local step improves the ``efficiency".

\subsection{Limitations}\label{app:limits}
While we present concrete and sound theoretical results, they heavily rely on Taylor's theorem, which only provides accurate information in the vicinity of the studied point. Thus, one might need to account for the impact of the error term once we start moving away from the studied point. Nevertheless, our experiments with finite step sizes, strongly support our theoretical analysis.

Indeed, the main point of our experiments is to show that our theoretical results carry on to practical settings. We do not claim, however, that our algorithm achieves state-of-the-art results, but sheds light on the impact that implicit regularization might have on the training of neural networks on non-artificial data sets. 

Both federated learning and distributed datacenter settings studied in this work heavily depend on many  hyperparameters (server momentum, normalization, learning rate decay scheduling, etc.) that we decided to ignore in this work. 
This allowed us to isolate the effect of implicit regularization, but it remains to study the interplay they have with FedGA. 

Lastly, the overhead in communication and computation cost in federated and distributed learning due to the calculation of the drift limits the scalability of our approach. 
Nevertheless, several techniques could be used to alleviate this issue~\citep{pmlr-v119-karimireddy20a}.

\subsection{Societal Impact}\label{app:impact}
We believe that collaborative learning schemes such as federated learning are an important element towards enabling privacy-preserving training of ML models, as well as for a better alignment of each participating individual's data ownership with the resulting utility from a jointly trained machine learning model, especially in applications where data is user-provided and privacy sensitive~\citep{kairouz2019advances,nedic2020review}.

In addition to privacy, efficiency gains in distributed training reduce the environmental impact of training large machine learning models.
The study of limitations of such methods in the realistic setting of heterogeneous data and algorithmic and practical improvements to the efficiency of such methods, is expected to help as a step towards achieving the goal of collaborative privacy-preserving and efficient decentralized learning.

\section{Experiments Appendix}
\subsection{Model architectures}\label{app:Architectures}
For the EMNIST experiments, we trained a CNN model with 2 convolutional layers followed by a fully connected layer. 

For the CIFAR10 experiments, we trained a CNN model with 2 convolutional layers followed by three fully connected layers. 

\subsection{Experiment Hyperparameters}\label{app:hyperparameters}

\subsubsection{Federated Learning}
We used a constant learning rate for each experiment, and we did not use momentum. 
For each algorithm we tuned the learning rate from \{0.05, 0.1, 0.2, 0.4\}. We tuned our algorithms with two batch sizes: 2400 corresponding to the entire dataset in each of the 47 workers, and 240 corresponding to 10\% of the worker's data. 
Weight decay ($L_2$ regularization) was tuned from \{0.001, 0.0001\}, where the former achieved better test accuracy in all reported cases.

The number of local steps of each algorithm was tuned from \{1, 10, 20, 40\}, which corresponds to 1, 10, 20, and 40 local epochs with batch-size 2400, respectively, and 0.1, 1, 2, and 4 local epochs with batch-size 240, respectively. In the IID setting, using batch-size 240 always achieved higher test accuracy. Furthermore, better generalization was achieved using either 10 local steps. Thus, the use of more local epochs might increase convergence speed in terms of the number of rounds, but has only a detrimental effect on the maximum test accuracy achievable; see Section~\ref{app:using_more_local_epochs}. 

The most challenging parameter to tune was $\beta$, the constant in front of the displacement in FedGA; see Algorithm~\ref{alg:FedGA}. 
We started with a coarse grid search with $\beta$ tuned from \{0.01, 0.1, 1.0, 5.0\}. After finding the best value in each of the two settings (IID and heterogeneous), we perform a fine grid search around it.
For the IID setting where the gradient variance is much smaller, we used a fine grid search with $\beta$ tuned from \{0.5, 1.5, 2.5, 3.5\}, with the best results for $\beta$ between 1.5 and 2.5.
In the heterogeneous setting, where the variance is much larger, we used a fine grid search with $\beta$ in \{0.01, 0.025 0.05, 0.1\}, with the best results between $0.025$ and $0.05$; orders of magnitude smaller than for the IID case. For more details, see Section~\ref{app:impact on tuning beta}.

\subsubsection{Datacenter distributed learning} 
We used a constant learning rate for each experiment, and we did not use momentum.
For each algorithm we tuned the learning rate from \{0.05, 0.1, 0.2, 0.4\}. 
Weight decay ($L_2$ regularization) was tuned from \{0.001, 0.0001\}, where the former achieved better Test Accuracy in all reported cases.

\paragraph{Sampling all clients}
Due to hardware and time constraints, we limited our search to a batch size of 125, which represents 2.5\% of the 5000 data examples in each worker.
The number of local steps of each algorithm was tuned within \{10, 20\}. In a federated learning setting, one might try to increase these numbers, but in the datacenter distributed setting, we assume that communication is not the bottleneck. Thus, while further experiments could be done, we believe our settings represent well the objectives of this paper. Furthermore, while we did not perform an exhaustive study for this task/architecture as in Section~\ref{app:using_more_local_epochs}, we also notice that an increase in local steps has not further benefited in test accuracy.

As in the federated learning setting, the most challenging parameter to tune was $\beta$.
We started with a coarse grid search with $\beta$ tuned from \{0.01, 0.1, 1.0, 5.0\}. After finding the best interval, we perform a fine grid search.
For this data center distributed setting, we found the gradient variance to be also quite small. Thus, we used a fine grid search with $\beta$ tuned from \{0.5, 1.5, 2.5, 3.5\}, with the best results for $\beta= 2.5$.

\paragraph{Minimizing number of updates.}
In these settings, we are restricted to using exactly one local step in each round.
We tuned our algorithms with two batch sizes: 1000 and 5000, corresponding to 20\% and 100\% of the worker's data, respectively.
The tuning of the $\beta$ parameter was performed in the same way as in the above setting, and the results were quite similar, with $\beta$ between 1.5 and 2.5 being the best range of values. 

\paragraph{Details of Hyperparameter Tuning}

For FedAvg, Scaffold, and LargeBatchSGD, we used an exhaustive grid search across all combinations of hyperparameters in our grid search. For example, for FedAvg we searched among the 5 learning rates, 2 weight decays, 2 batch-sizes, and 4 numbers of local steps, i.e., a total of 80 experiments. During training, we terminated the experiments with a significantly lower validation accuracy after $20\%$ of the rounds. For FedGA and GradAlign, we transferred the optimal hyperparameters for FedAvg and LargeBatch SGD and fixed them to then perform the search for the best parameter beta, as this depends on the variance and smoothness according to our theoretical analysis. Once the best value of beta was determined, we fixed beta and performed the same exhaustive hyperparameter search over the other aforementioned hyperparameters. That is, due to computational constraints, beta was the only parameter tuned independently. Since we obtained positive results already with this method, we did not pursue further improvements.
\subsection{Tuning the $\beta$ parameter of FedGA}\label{app:impact on tuning beta}
\begin{figure}[ht]
\centering
\includegraphics[width=0.49\textwidth]{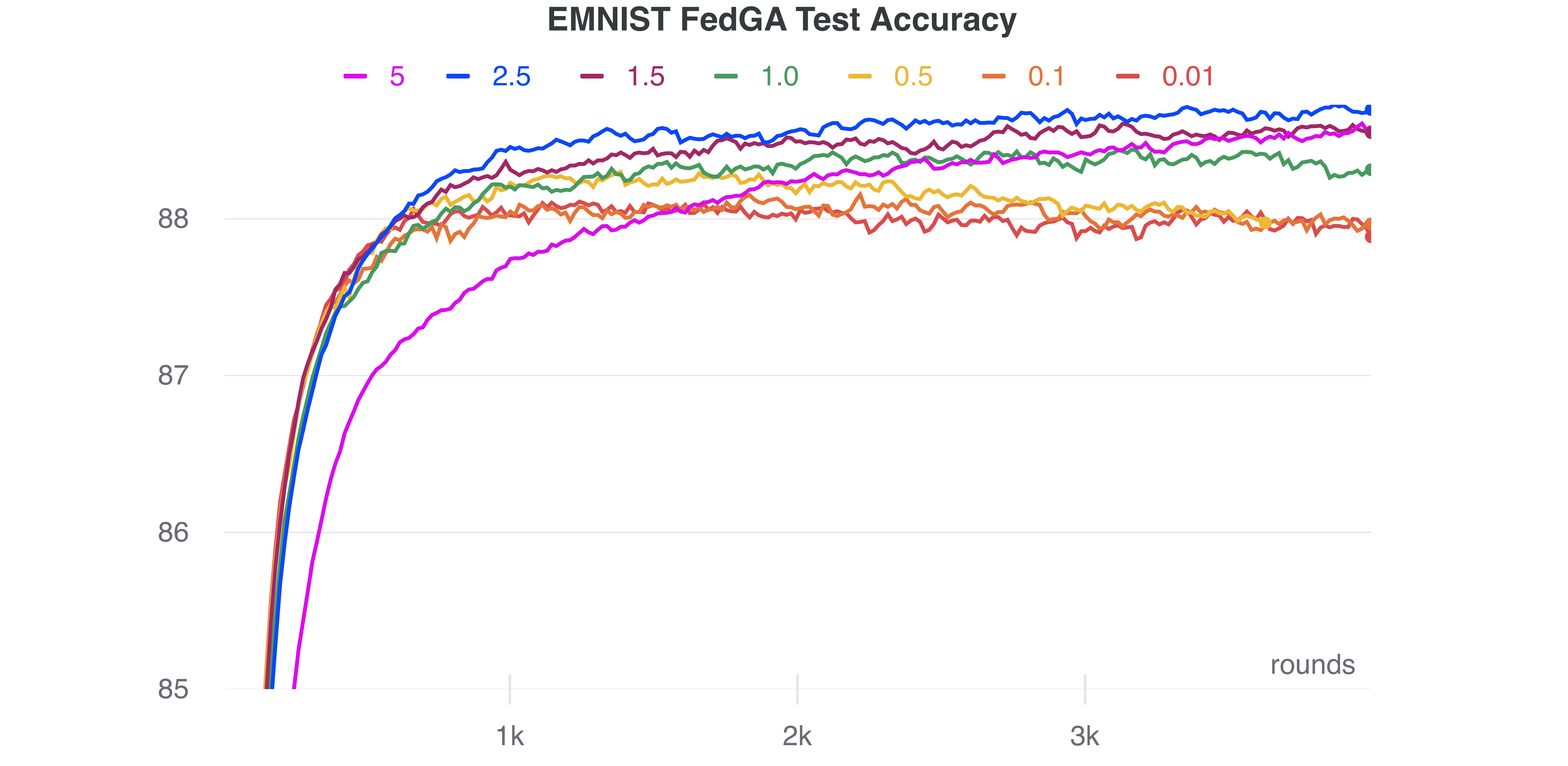}
\includegraphics[width=0.49\textwidth]{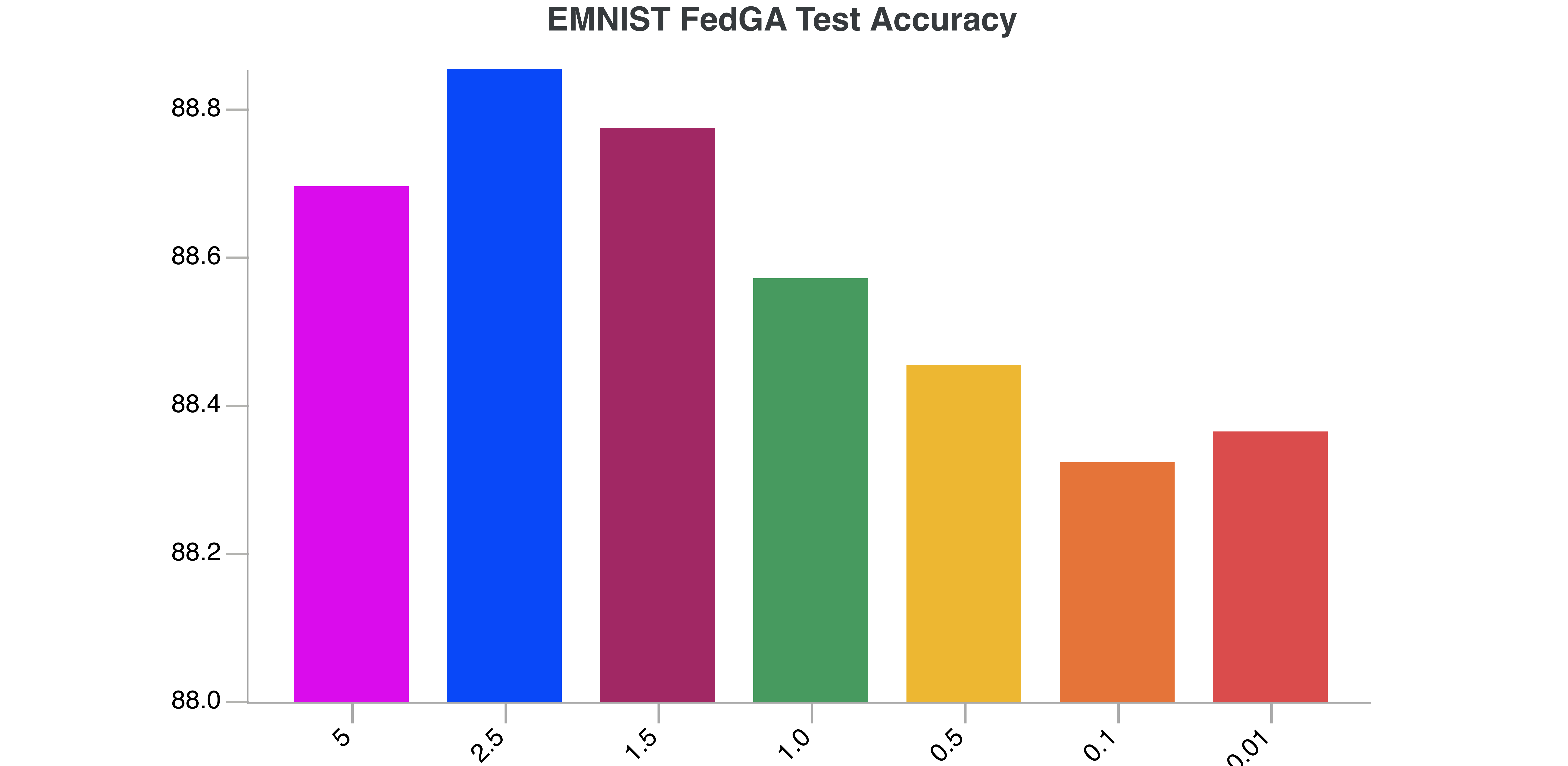}
\caption{\small Depicts the effect of tuning the parameter $\beta$ for one of the gird search settings we tried. For this example, we fix the batch size to 240 in the IID setting, weight decay 0.001, learning rate 0.2, and 10 local steps.
From our experiments, it seems that the test accuracy as a function of beta is concave, which might help with its optimization. The experiments were performed with the same initial random seed.}
\label{fig:beta tuning}
\end{figure}

As mentioned in Section~\ref{app:hyperparameters}, it was challenging to tune $\beta$ as it depends heavily on the variance of the gradients. 
In our experiments, we used a coarse level grid search, follower by a fine-tuning.
However, as depicted in Figure~\ref{fig:beta tuning}, it might seem that the test accuracy as a function of beta might be a concave function, which can greatly help with its optimization. 
While the possibility of modifying $\beta$ seems to offer an advantage over SCAFFOLD, it brings along the additional challenge of tuning it.

\subsection{Effect of local epochs}\label{app:using_more_local_epochs}
\begin{figure}[ht]
\centering
\includegraphics[width=0.49\textwidth]{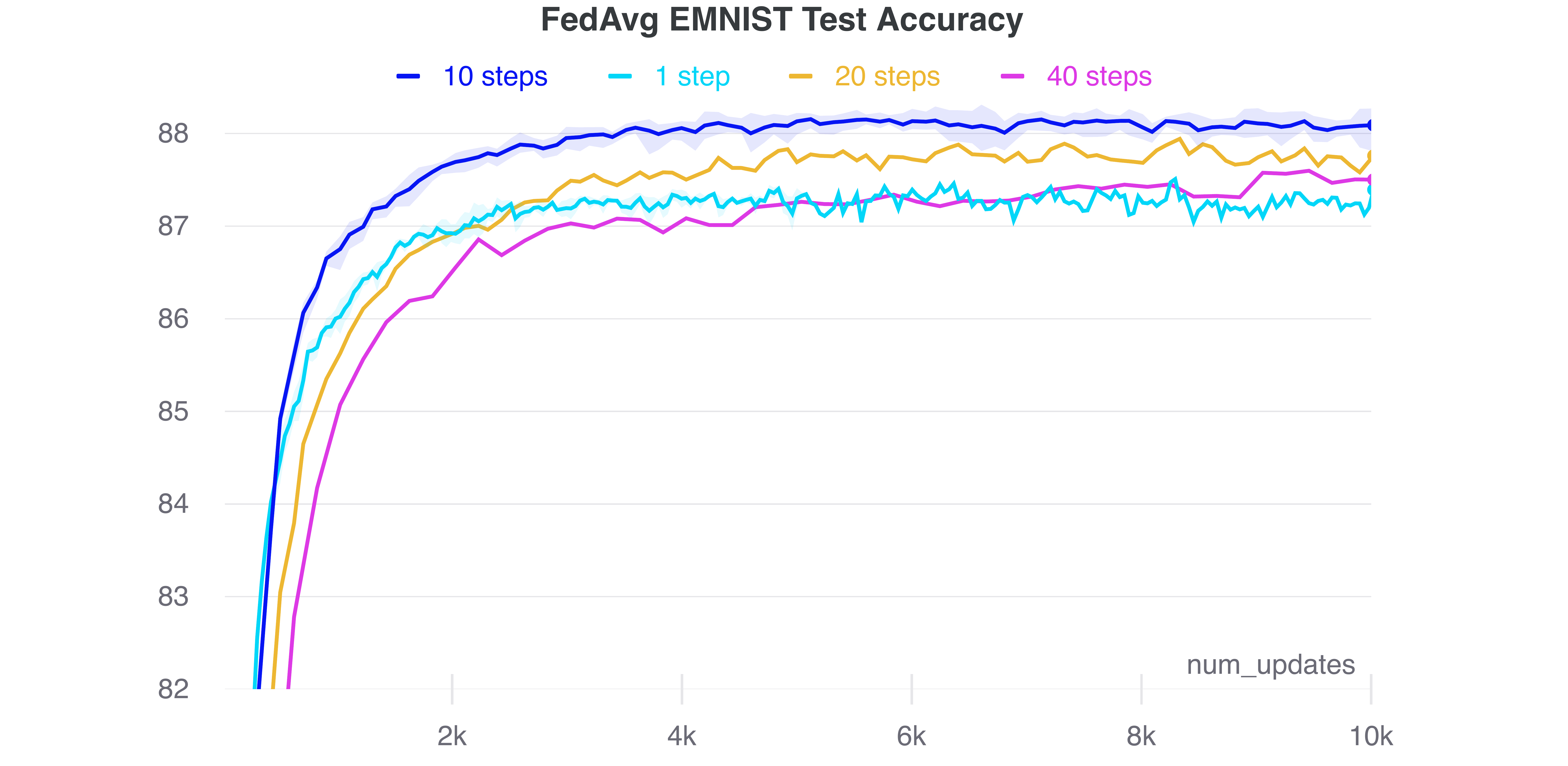}
\includegraphics[width=0.49\textwidth]{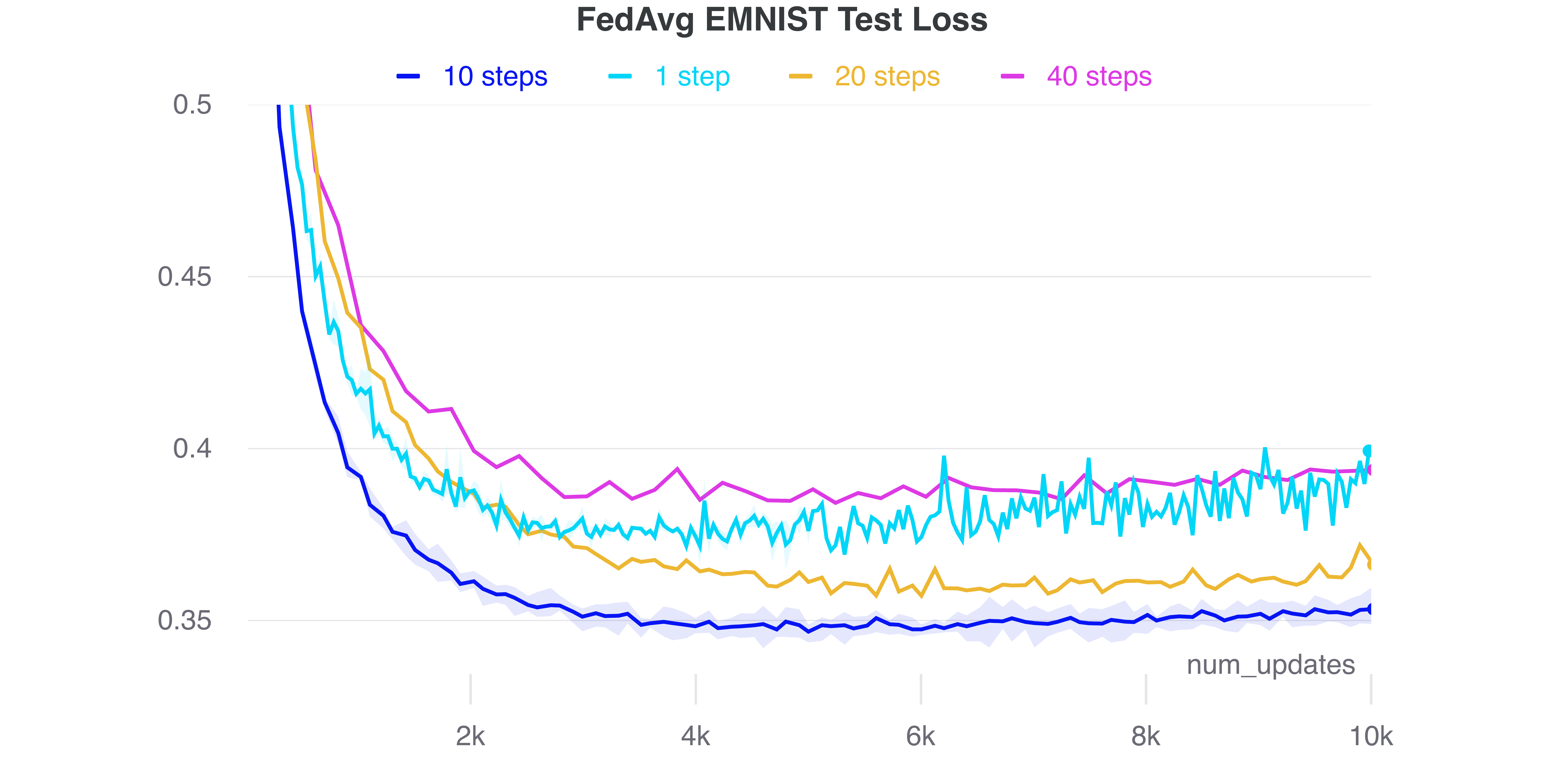}
\caption{\small Best performances of FedAvg with batch size 240 and IID data distribution within our grid search. The $x$-axis shows the total number of (local) updates performed by the algorithm.}
\label{fig:more steps are not better}
\end{figure}

For our grid-search with the number of steps in \{1,10,20,40\}, which corresponds to 0.1, 1, 2, and 4 local epochs, we notice that beyond 10 local steps, there is no generalization benefit. Moreover, we can see a detriment in the maximum test accuracy; see Figure~\ref{fig:more steps are not better}. There is, however, a much faster convergence using more local steps, but to a model with worse test accuracy. Similar behavior was spotted in FedGA and Scaffold.
While this phenomenon might be overcome by a further reduction of the learning rate, this was beyond the parameters in our grid search. 

\subsection{Results for NLP tasks}\label{app:NLP results}
We also investigate the task of next-character prediction on the dataset of ``The Complete Works of William Shakespeare'' \citep{mcmahan2017communication}. Each speaking role in the plays is assigned to a different client. 
We take a small subsample of this dataset using the LEAF partitioning script~\cite{caldas2018leaf} to obtain 134 workers with a total of 420,117 samples and an average of 3135 samples per worker.
We use a two-layer LSTM classifier containing 100 hidden units with an 8-dimensional embedding layer. We used a sequence length of 80 and the task is to predict the next character. 

\begin{figure}[h!]
\centering
\includegraphics[width=0.49\textwidth]{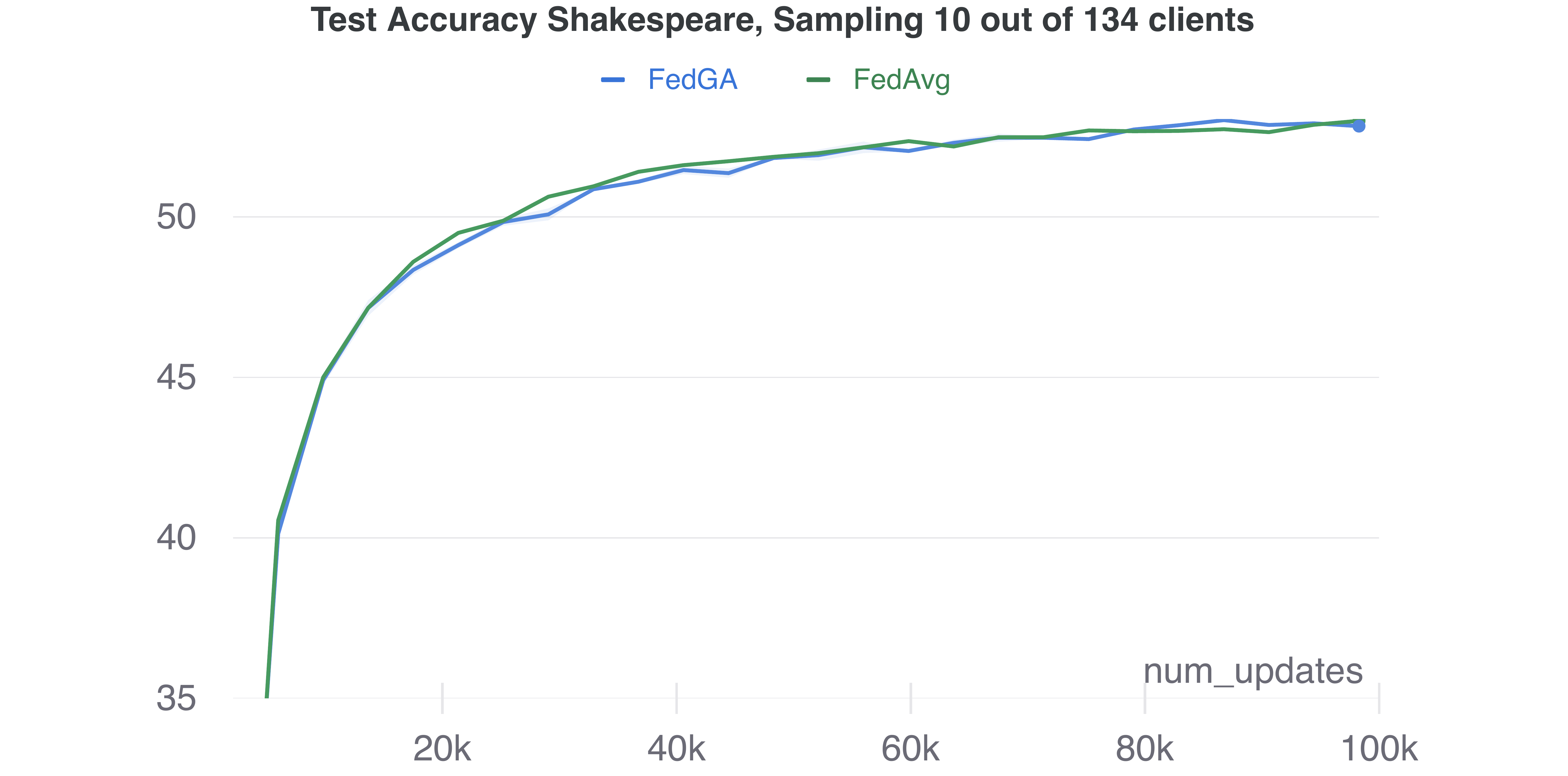}
\includegraphics[width=0.49\textwidth]{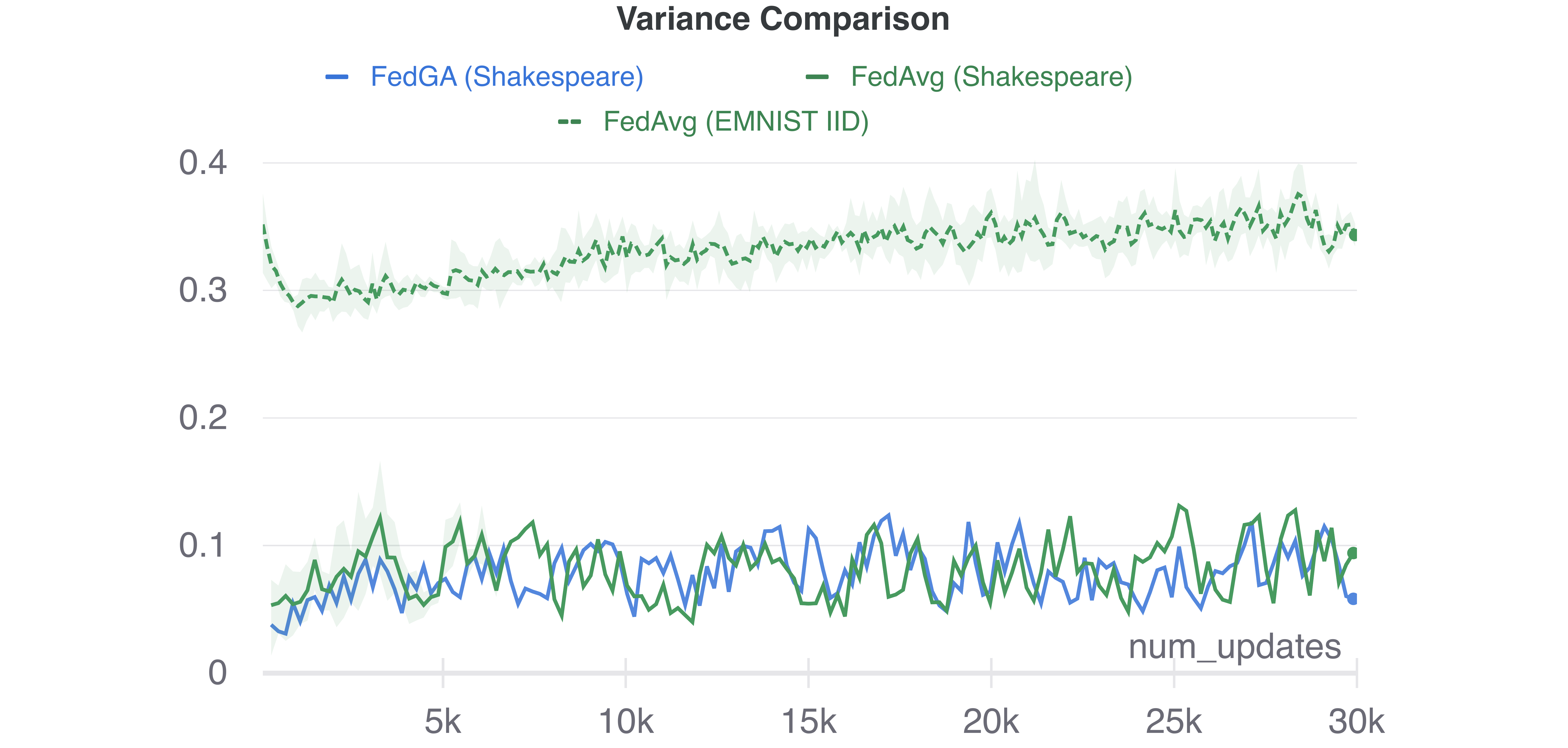}
\caption{\small Experiments on the Shakespeare dataset using an LSTM RNN architecture for the federated learning setting. Left: The plot shows that in terms of updates, FedGA and FedAvg are equivalent in terms of test accuracy. Right: A comparison of the magnitude of the difference between the global objective gradient and the gradient of the first client's objective, i.e.,  $\|\nabla f(\xx) - \nabla f_i(\xx)\|$. FedGA and FedAvg have a very low variance for the Shakespeare dataset, even with the non-iid distribution. In contrast, the variance for FedAvg  in the IID setting is much larger.}
\label{fig:Shakespeare}
\end{figure}

For this experiment, however, we could not obtain an advantage while using FedGA. Regardless of the $\beta$ used during our grid search, the performance of FedGA was equivalent to that of FedAvg in terms of the number of updates. To explain this, we took a closer look at the gradient alignment (see Figure~\ref{fig:Shakespeare}), where we realize that the magnitude of the difference between the global objective gradient and the gradient of the first client's objective, i.e.,  $\|\nabla f(\xx) - \nabla f_i(\xx)\|$, is very small. That means that gradients are already aligned between workers, and hence additional gradient alignment does not help.

This behavior can be explained by the bias of the English language in the next-character prediction distribution. Even if the style of each character is slightly different, most of the words they use are the same, and the spelling of these words is the same among all the workers. This leads to a very small variance among the gradients. We will further study NLP tasks on datasets with less gradient alignment.

\subsection{Results for CIFAR100}\label{app:CIFAR100}
We performed additional experiments in the federated learning setting with the CIFAR100 dataset~\citep{krizhevsky2009learning} consisting of 60,000 $32\times32$ images, 600 per class. We used the heterogeneous distribution and used 50 clients, i.e., each client contains images from exactly 2 classes, and no two clients share images of the same class. We sample 10 clients in each round.

Our model and grid search are identical to those used for the CIFAR10 experiments, with the exception of use of gradient clipping to max norm 5 which helped to stabilize the training. Within our hyperparameter search, SCAFFOLD failed to train to a better than random accuracy, even with the use of gradient clipping.

\begin{figure}[h!]
\centering
\includegraphics[width=0.49\textwidth]{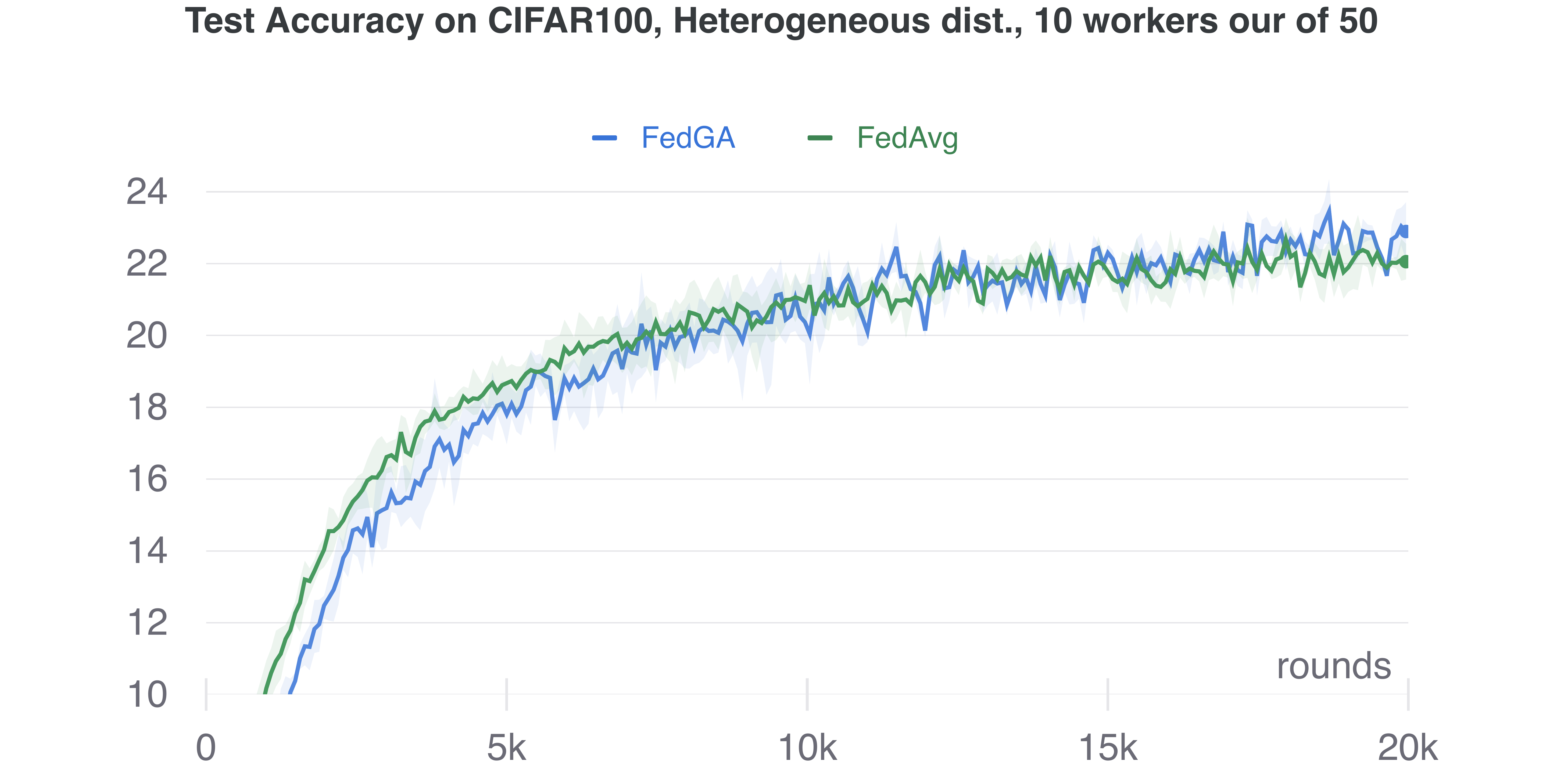}
\includegraphics[width=0.49\textwidth]{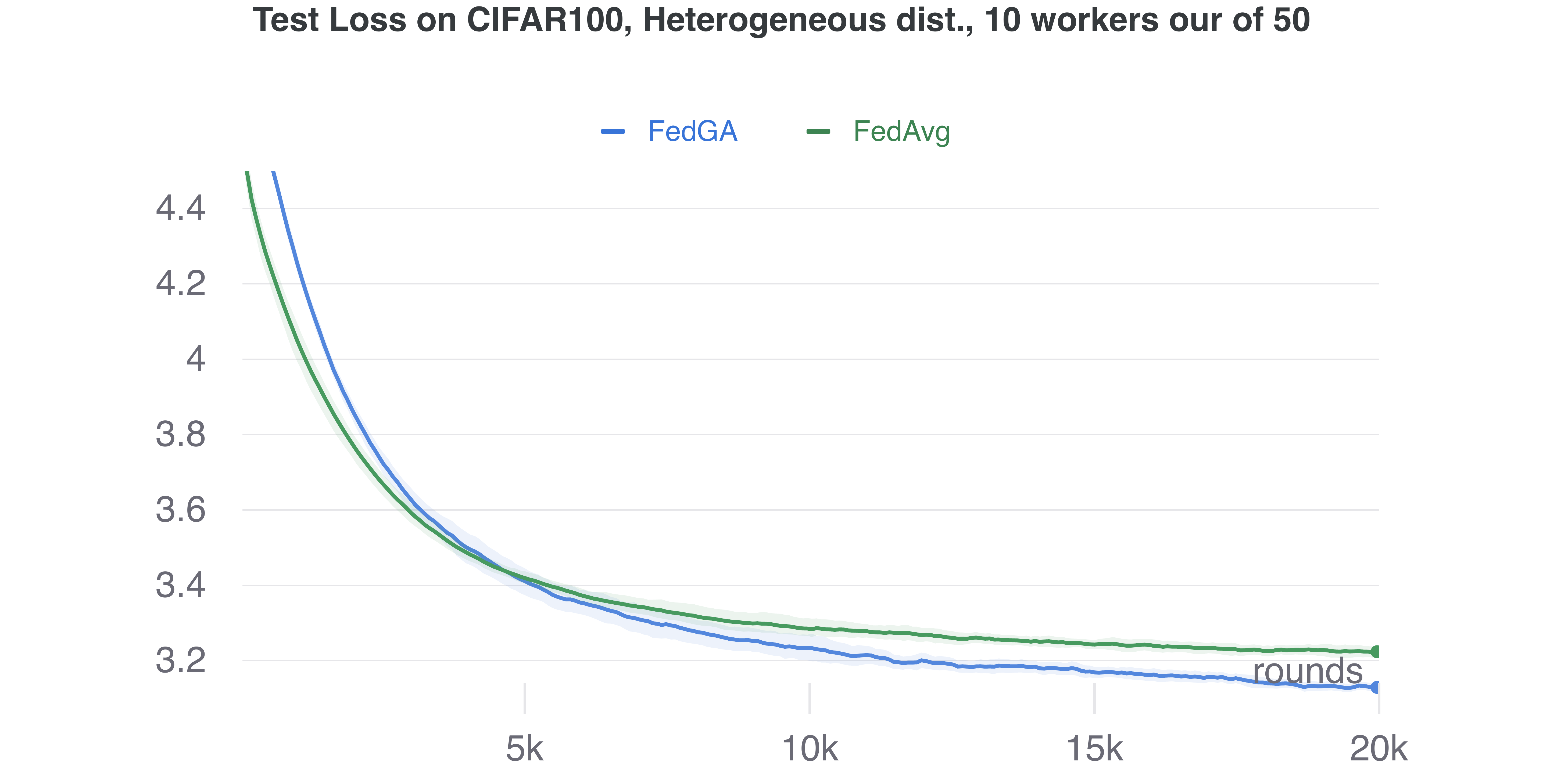}
\caption{\small Experiments on the CIFAR100 dataset using a CNN architecture for the federated learning setting with 50 clients, out of which 10 are uniformly sampled in every round.}
\label{fig:CIFAR100 results}
\end{figure}

\begin{figure}[ht]
\centering
\includegraphics[width=0.49\textwidth]{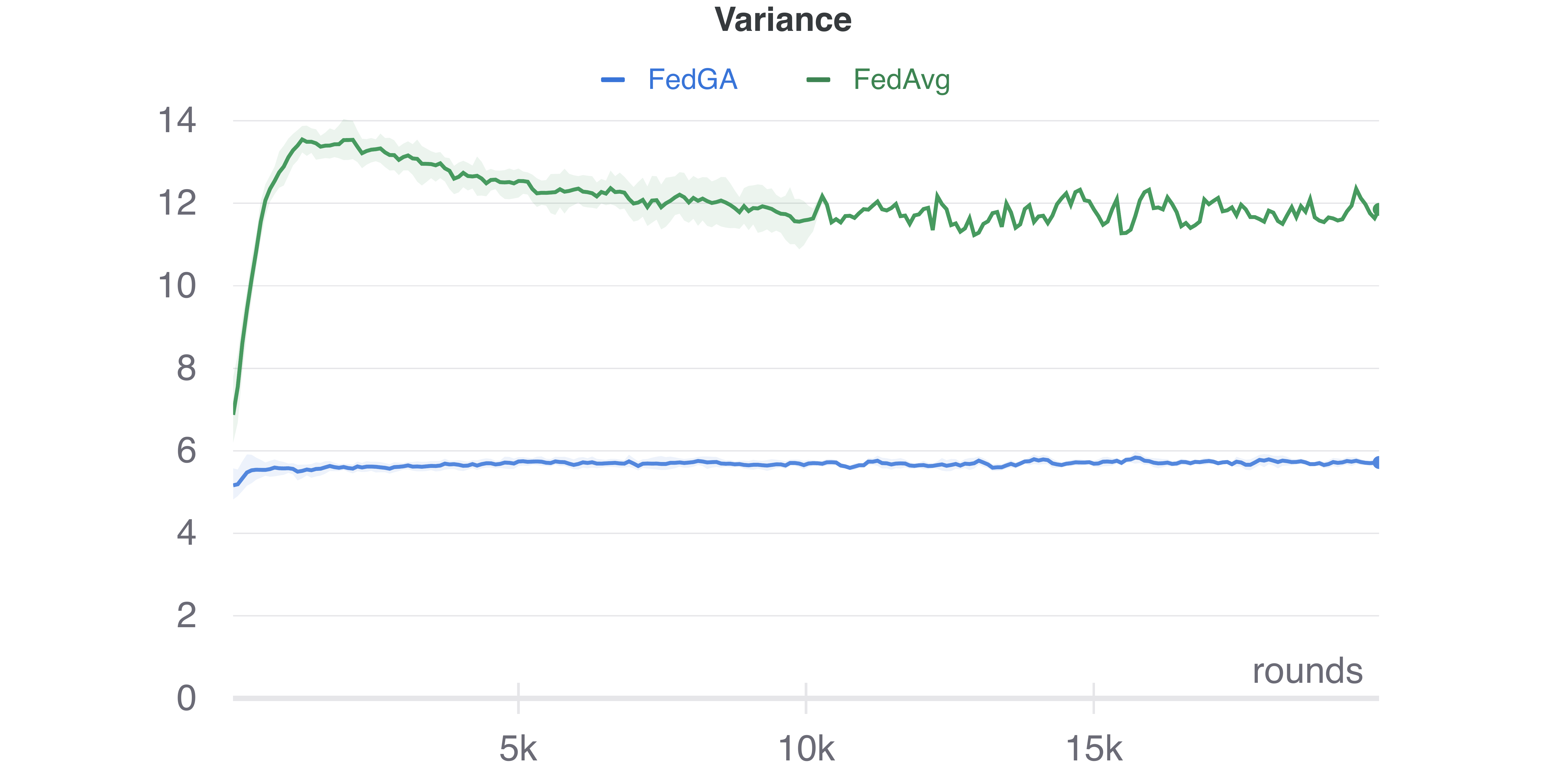}
\caption{\small Magnitude of the difference between the global objective gradient and the gradient of the first client's objective, i.e.,  $\|\nabla f(\xx) - \nabla f_i(\xx)\|$, over 20000 rounds of training on the CIFAR100 dataset. FedGA achieves a clear variance reduction during training that translates into a higher final Test Accuracy and significantly smaller Test Loss.}
\label{fig:CIFAR100 Variance}
\end{figure}

As shown in Figure~\ref{fig:CIFAR100 results}, FedGA achieves a higher Test Accuracy of $24.09 \pm 0.54$ while FedAvg achieves $23.63 \pm 0.43$ after averaging over 3 runs. The test loss in the right of the figure reflects more clearly the advantage of using FedGA. As always, the plots include the extra rodund of communication used by FedGA. Figure~\ref{fig:CIFAR100 Variance} shows a clear alignment in the gradients of FedGA in comparison to those of FedAvg.

\subsection{Training Accuracy}\label{app:training accuracy}
\begin{figure}[ht]
\centering
\includegraphics[width=0.49\textwidth]{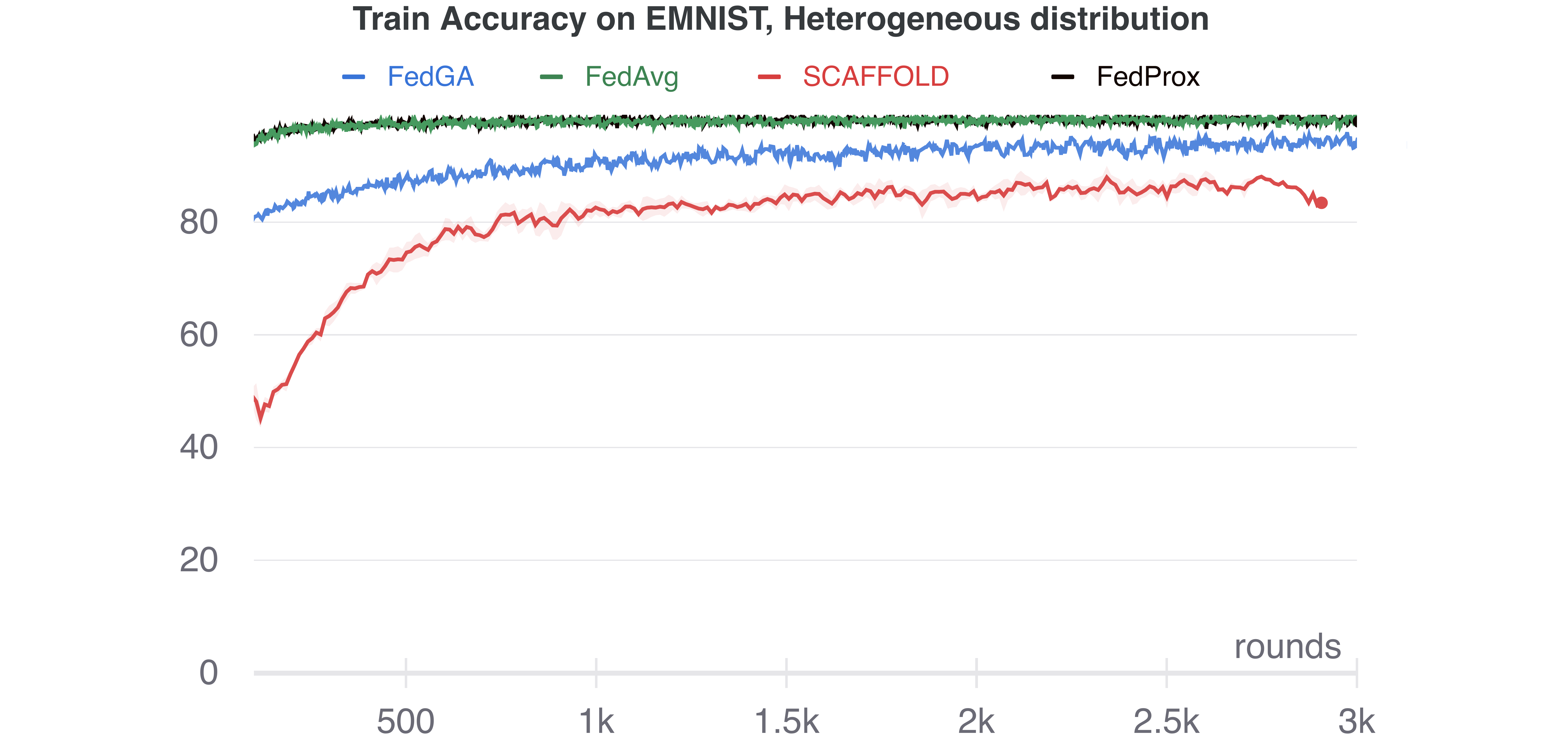}
\includegraphics[width=0.49\textwidth]{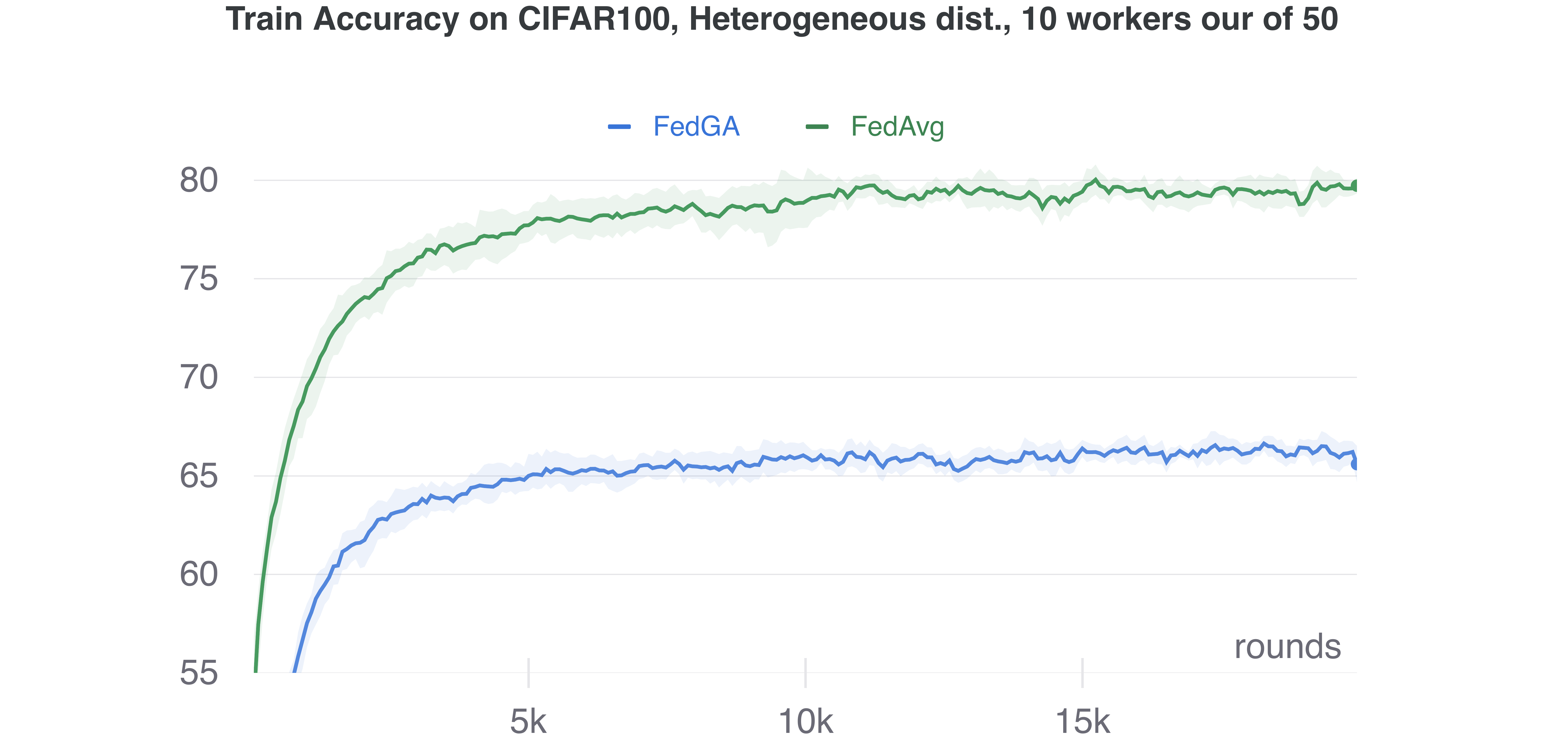}
\caption{\small Train accuracies for both EMNIST (left) and CIDAR100 (right). The regularization effect of both SCAFFOLD and FedGA leads to less overfitting on the training data.}
\label{fig:TrainAccuracies}
\end{figure}

Figure~\ref{fig:TrainAccuracies} Shows the train accuracies for both the training of CIFAR100 and EMNIST with a heterogeneous data distribution. One can clearly see the regularization effect of FedGA which leads to a much lower train accuracy, but to a higher test accuracy.

\subsection{Additional Plots}\label{app:plots}
\begin{figure}[t]
\centering
\includegraphics[width=0.49\textwidth]{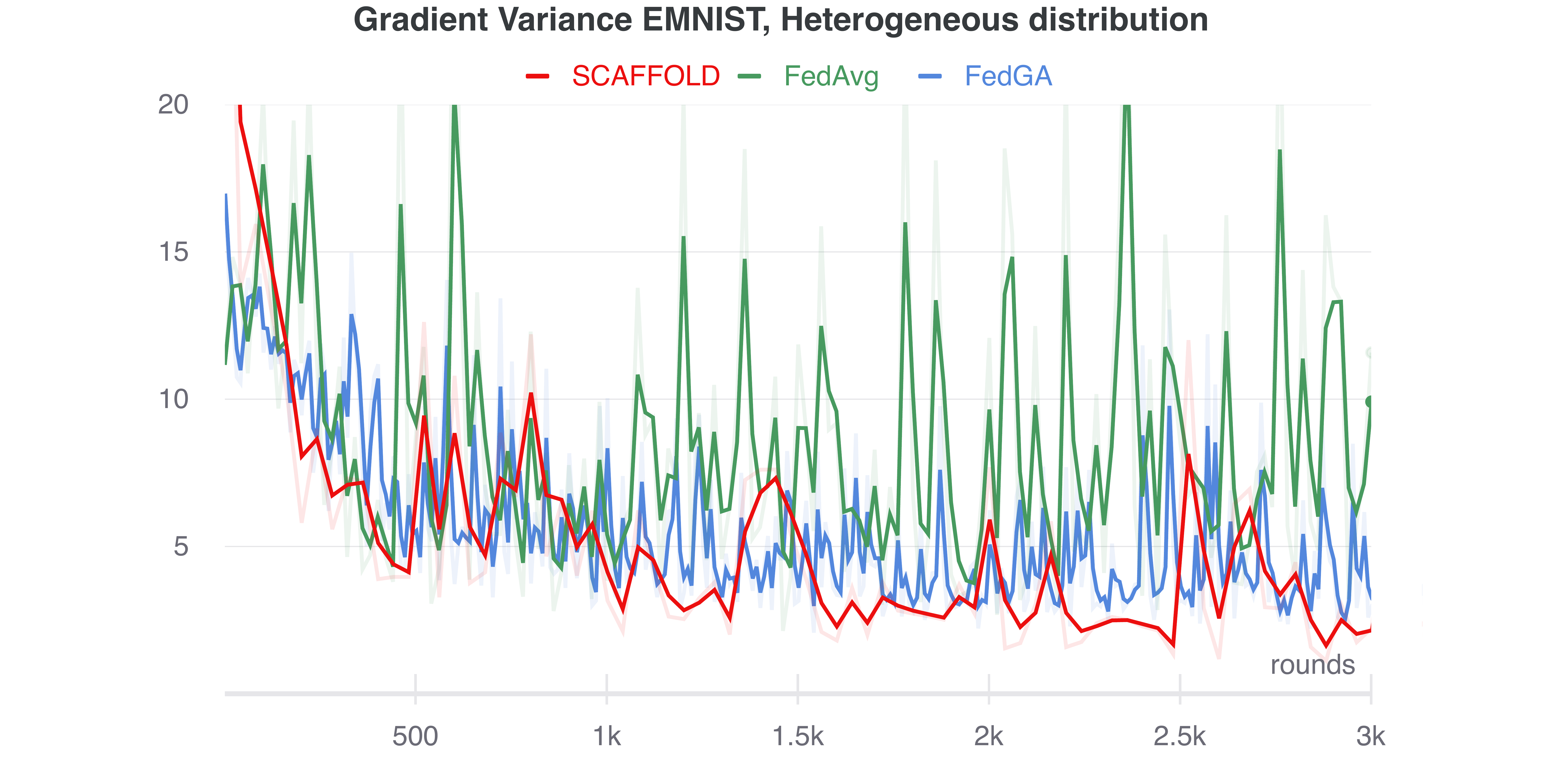}
\includegraphics[width=0.49\textwidth]{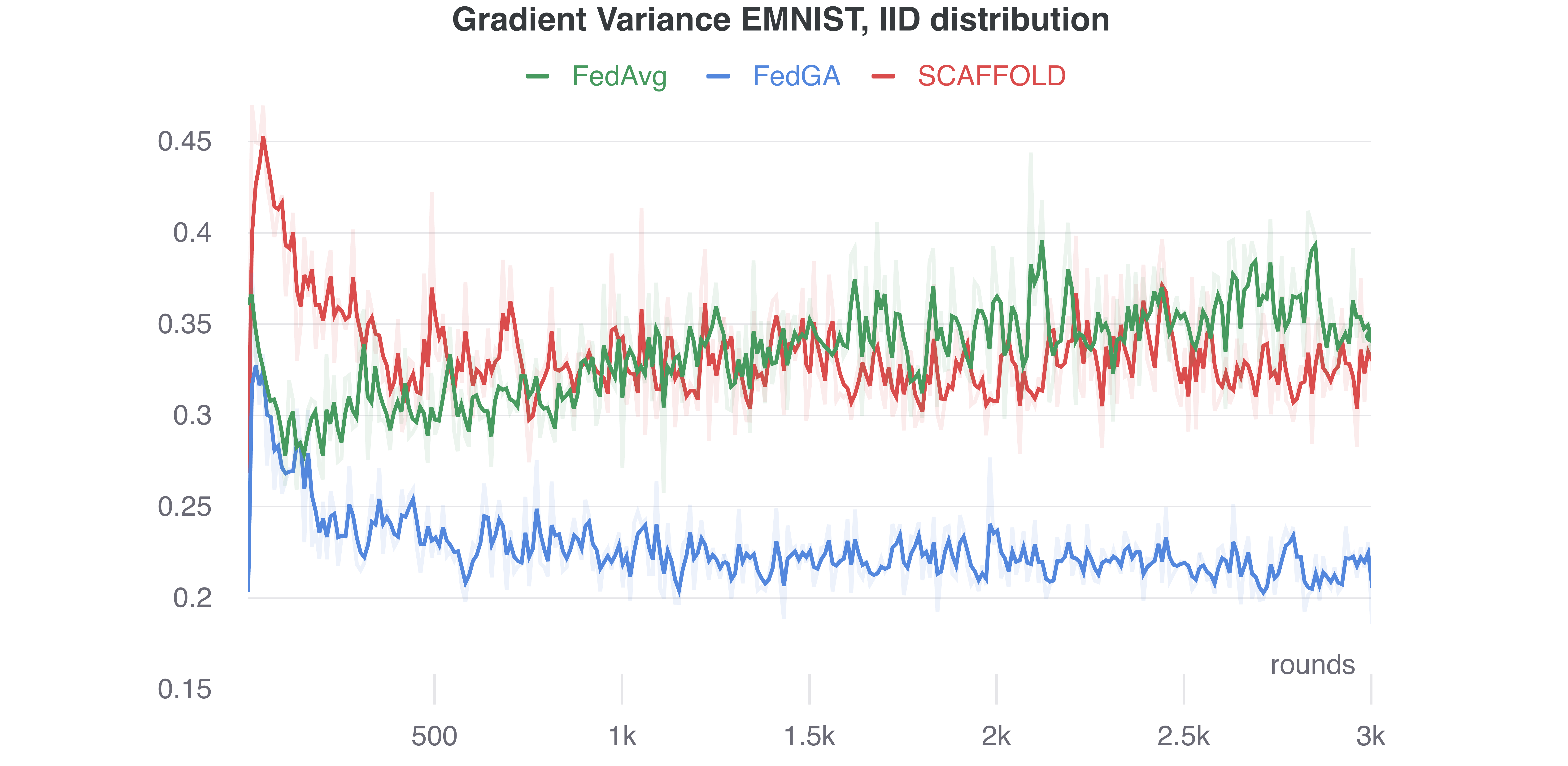}
\vspace{-3mm}
\caption{\small Magnitude of the difference between the global objective gradient and the gradient of the first client's objective, i.e.,  $\|\nabla f(\xx) - \nabla f_i(\xx)\|$, over 3000 rounds of training.
The magnitude of this difference is much smaller in the IID- than in the heterogeneous setting. However, in both cases, FedGA and SCAFFOLD tend to have a smaller difference than FedAvg. 
Moreover, not only is this quantity lower, but it has a smaller variability.
}\vspace{-2mm}
\label{fig:gradient_variance}

\end{figure}

\begin{figure}[h]
\centering
\includegraphics[width=0.49\textwidth]{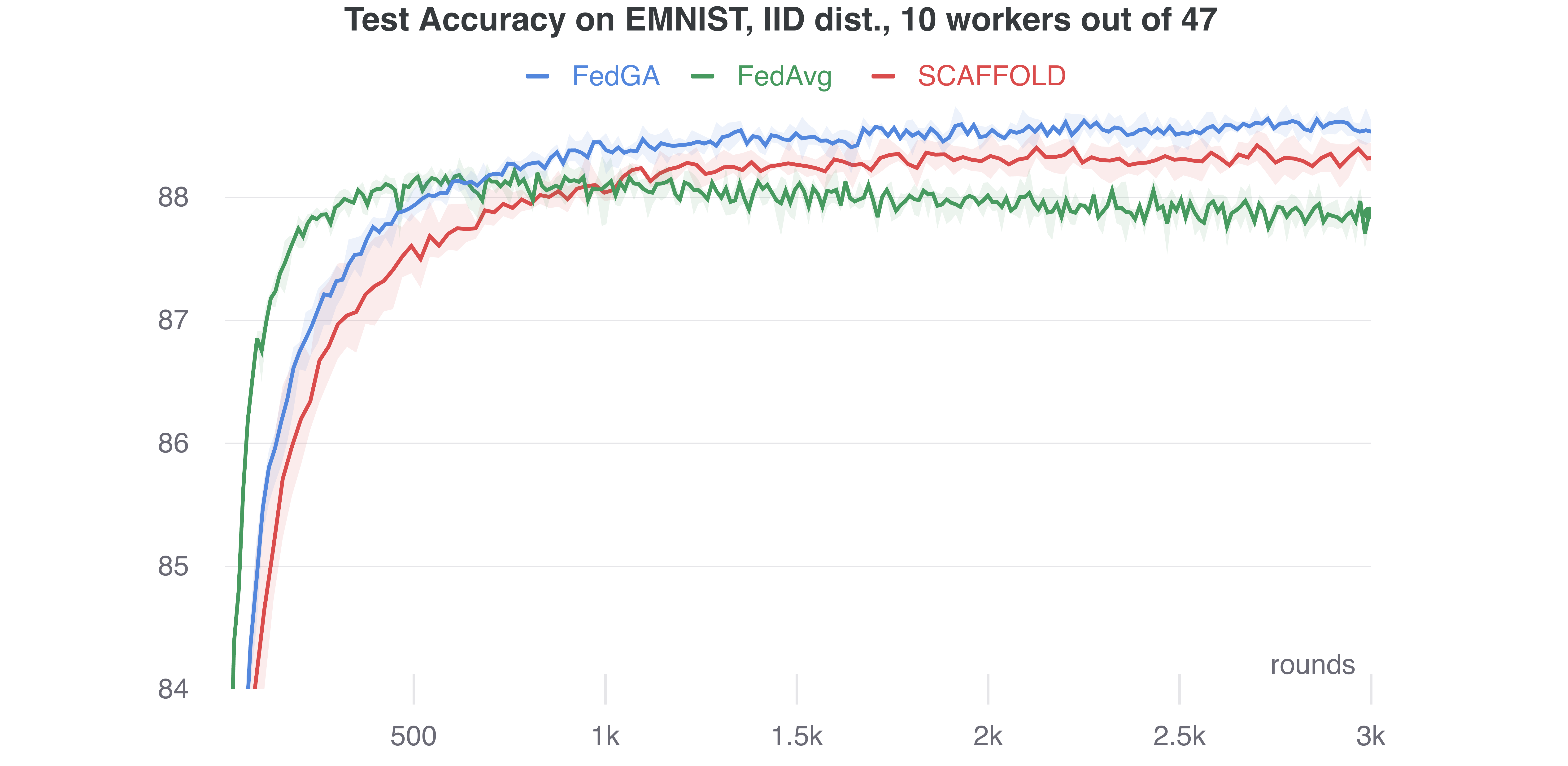}
\includegraphics[width=0.49\textwidth]{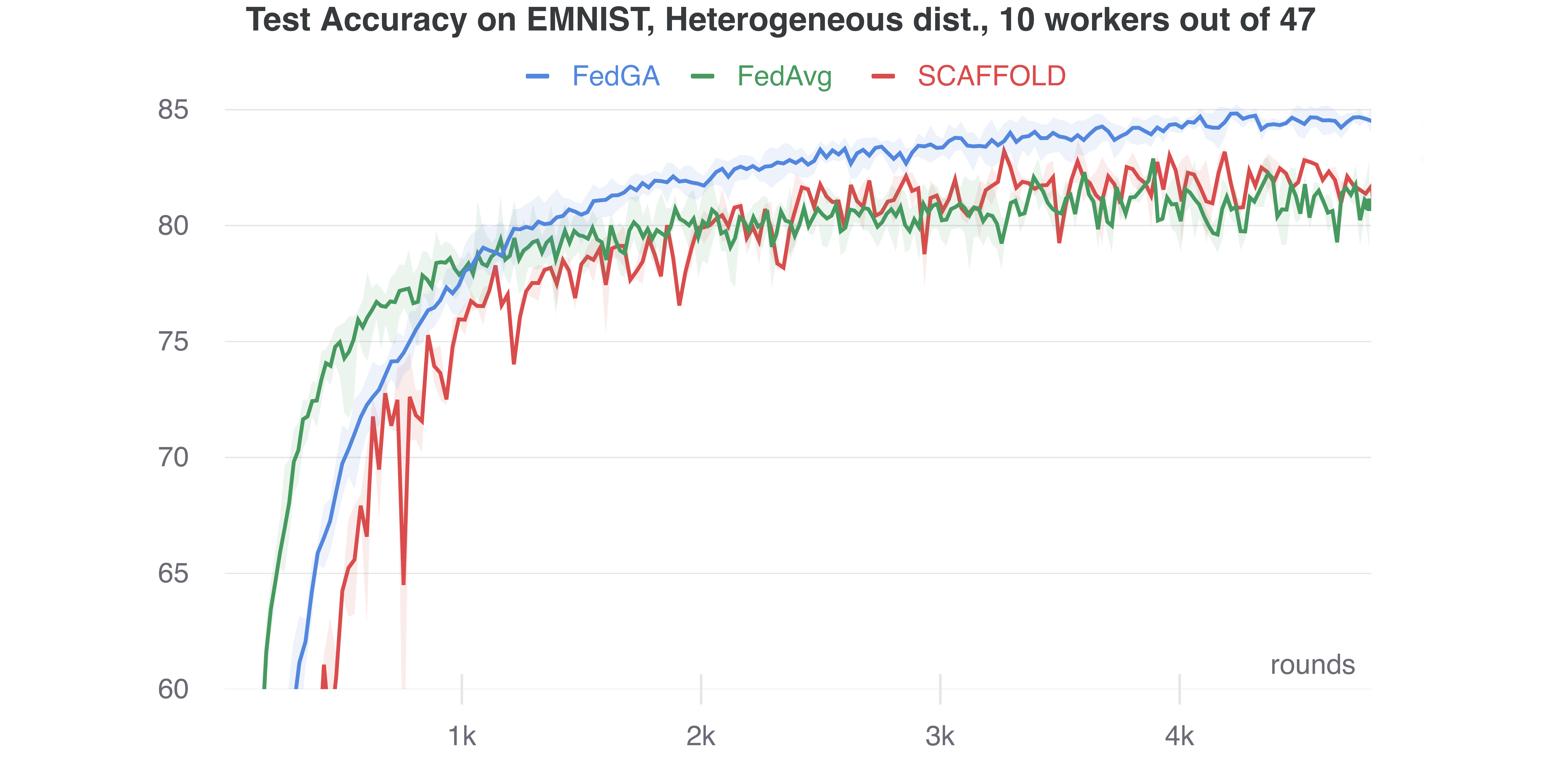}
\vspace{-3mm}
\caption{\small 
Experiments on the EMNIST dataset using a CNN architecture for the federated learning setting with 47 clients, out of which 10 are uniformly sampled in every round. While FedAvg is faster and can efficiently use more local epochs, both FedGA and SCAFFOLD generalize better.
Left: Data is distributed using the IID setting, where data for each client is drawn uniformly at random. 
Right: Data is distributed  heterogeneously, each client having examples of only a single class. 
This is the most challenging setting for federated algorithms.}
\label{fig:emnist_iid}
\end{figure}

Figures~\ref{fig:gradient_variance} and~\ref{fig:emnist_iid} illustrate a comparison between FedGA, FedAvg and SCAFFOLD for the EMNIST dataset. They complement the results presented in the main body of the paper.

\end{document}